\documentclass[10pt,twocolumn,letterpaper]{article}

\usepackage{iccv}
\usepackage{times}
\usepackage{epsfig}
\usepackage{graphicx}
\usepackage{amsmath}
\usepackage{amssymb}
\usepackage[pagebackref,breaklinks,colorlinks]{hyperref}
\usepackage{mathtools}
\usepackage{amsthm}
\usepackage{multirow}
\usepackage{booktabs}
\usepackage{bbm}
\usepackage{amsmath}
\usepackage{amssymb}
\usepackage{mathtools}
\usepackage{amsthm}
\usepackage{multirow}
\usepackage{multicol}
\usepackage{microtype}
\usepackage{graphicx}
\usepackage{subfigure}
\usepackage{booktabs}
\theoremstyle{plain}

\theoremstyle{definition}

\theoremstyle{remark}

\def\b{\mathbf{b}}
\def\r{\mathbf{r}}

\def\u{\mathbf{u}}
\def\v{\mathbf{v}}
\def\h{\mathbf{h}}
\def\m{\mathbf{m}}

\def\x{\mathbf{x}}
\def\q{\mathbf{q}}

\def\R{\mathbb{R}}

\def\model{\text{IIA }}

\iccvfinalcopy %

\def\iccvPaperID{12462} %

\ificcvfinal\fi

\begin{document}

\title{Visual Explanations via Iterated Integrated Attributions}

\author{\vspace{2mm}Oren Barkan\thanks{Denotes equal contribution.}$^{\hspace{1.5mm}1}$
\hspace{10mm} 
Yehonatan Elisha$^{\ast \hspace{0.2mm} 1}$
\hspace{10mm}
Yuval Asher$^{2}$\\
\vspace{3mm}Amit Eshel$^{2}$
\hspace{10mm}
Noam Koenigstein$^{2}$ 
\\
\vspace{2mm}
\normalsize{$^{1}$The Open University}
\hspace{10mm} \normalsize{$^{2}$Tel Aviv University}
\\
\normalsize{\url{https://github.com/iia-iccv23/iia}}
}

\maketitle
\ificcvfinal\thispagestyle{empty}\fi

\begin{abstract}

We introduce Iterated Integrated Attributions (IIA) - a generic method for explaining the predictions of vision models. IIA employs iterative integration across the input image, the internal representations generated by the model, and their gradients, yielding precise and focused explanation maps. We demonstrate the effectiveness of IIA through comprehensive evaluations across various tasks, datasets, and network architectures. Our results showcase that IIA produces accurate explanation maps, outperforming other state-of-the-art explanation techniques.
\end{abstract}

 \section{Introduction}
\label{sec:intro}

The emergence of deep learning has ushered in significant breakthroughs within the realm of artificial intelligence, particularly in computer vision. Advanced deep Convolutional Neural Networks (CNNs) architectures~\cite{simonyan2014very, He2016DeepRL, Huang2017DenselyCC, Liu2022ACF}, and recent Vision Transformer (ViT) models~\cite{dosovitskiy2020image_vit,he2022masked_mae} have demonstrated state-of-the-art performance in image classification~\cite{NIPS2012_c399862d,simonyan2014very}, object detection~\cite{he2017mask,dai2016r,carion2020end_detr}, and semantic segmentation ~\cite{dai2016r, badrinarayanan2017segnet} tasks. Yet, many deep learning models lack interpretability, making it difficult to explain the reasoning behind their predictions. As a result, Explainable AI (XAI) has become a prominent research area in computer vision, and numerous methods have been proposed for explaining and interpreting the internal workings of different neural network architectures in various application domains~\cite{zeiler2014visualizing,simonyan2013deep,selvaraju2017grad, chefer2021transformer, barkan2021grad, malkiel2022interpreting,barkan2020explainable,barkan2023deep,barkan2023ltx,barkan2023six,gaiger2023not}.

Explanation methods attempt to produce an explanation map in the form of a heatmap (also known as \emph{relevance} or \emph{saliency} map) that attributes the prediction to the input by highlighting specific regions in the input image. Early gradient-based methods produced explanation maps based on the gradient of the prediction w.r.t. the input image ~\cite{simonyan2013deep,simonyan2014very,springenberg2014striving}. Then, Grad-CAM~\cite{selvaraju2017grad} and the follow-up works by~\cite{chattopadhay2018grad,jiang2021layercam, barkan2021gam} proposed to compute the explanation maps based on the internal activation maps (also known as Class Activation Maps (CAM)) and their corresponding gradients. In parallel, path integration methods such as Integrated Gradients (IG)~\cite{SundararajanTY17} proposed to produce an explanation map by accumulating the gradients of the linear interpolations between the input and reference images. The aforementioned techniques were formulated and evaluated on CNNs. Following the advent of Transformer-based architectures~\cite{vaswani2017attention}, a variety of approaches has also been proposed for interpreting Vision Transformer (ViT) models~\cite{chefer2021transformer,voita2019analyzing, chefer2021generic}.

This paper presents Iterated Integrated Attributions (IIA) - a universal technique for explaining vision models, applicable to both CNN and ViT architectures. IIA employs iterative integration across the input image, the internal representations generated by the model, and their gradients. Thereby, IIA leverages information from the activation (or attention) maps created by all network layers, including those from the input image. We present comprehensive objective and subjective evaluations that demonstrate the effectiveness of IIA in generating faithful explanations for both CNN and ViT models. Our results show that IIA outperforms current state-of-the-art methods on various explanation and segmentation tests across all datasets, model architectures, and metrics.

\section{Related Work}
\label{sec:related}
\paragraph{Explaining CNNs} Explanation methods for CNNs have been studied extensively. Saliency-based methods~\cite{dabkowski2017real,simonyan2013deep,mahendran2016visualizing,zhou2016learning,zeiler2014visualizing,zhou2018interpreting} and activation-based methods~\cite{erhan2009visualizing} use the feature-maps obtained by forward propagation in order to interpret the output prediction.
Perturbation-based methods~\cite{fong2019understanding,fong2017interpretable}
measures the output's sensitivity w.r.t. the input using random perturbations applied in the input space.
Gradient methods produce explanation maps based on the gradient itself or via a function that combines the activation maps with their gradients~\cite{Shrikumar2016NotJA, Srinivas2019FullGradientRF}. A prominent example is the Grad-CAM (GC)~\cite{selvaraju2017grad} method that uses the pooled gradients and the activation maps to produce explanation maps. GC attracted much attention from the XAI community with several follow-up works~\cite{chattopadhay2018grad,Fu2020AxiombasedGT,jiang2021layercam,barkan2021gam}. 
Another relevant line of work is path integration methods. Integrated Gradients (IG)~\cite{SundararajanTY17} integrates over the interpolated image gradients. Blur IG (BIG)~\cite{xu2020attribution} is concerned with the introduction of information using a baseline and opts to use a path that progressively removes Gaussian blur from the attributed image. Guided IG (GIG)~\cite{kapishnikov2021guided} improves upon IG by introducing the idea of an adaptive path method. By calculating the integration along a different path, high gradient areas are avoided which often leads to an overall reduction in irrelevant attributions. 
Differing from IG, GIG, and BIG, IIA employs iterated integration, enabling interpolation of the complete set of activation (attention) maps across all network layers. Moreover, IIA does not limit the integrand to plain gradients but accommodates any arbitrary function involving the activation (attention) maps and their gradients..
Gradient-free methods produce explanation maps via manipulation over the activation maps without relying on gradient information~\cite{Wang2020ScoreCAMSV,Desai2020AblationCAMVE}. For instance, LIFT-CAM employs the DeepLIFT~\cite{Shrikumar2017LearningIF} technique to estimate the activation maps' SHAP values~\cite{Lundberg2017AUA}, which are then combined with the activation maps to produce the explanation map. However, since these methods do not consider gradient information, their ability to effectively guide explanations towards the predicted class is limited.

\vspace{-5mm}
\paragraph{Explaining ViTs} Initial attempts to interpret Transformers utilized the inherent attention scores of ViT models to gain insights into input processing~\cite{vaswani2017attention,carion2020end}. However, the challenge lay in effectively combining scores from different layers. Simple averaging of attention scores for each token, for instance, often resulted in signal blurring~\cite{abnar2020quantifying,chefer2021transformer}. Abnar and Zuidema introduced the Rollout method, which computes attention scores for input tokens at each layer by considering raw attention scores within a layer as well as those from preceding layers~\cite{abnar2020quantifying}. Rollout showed improvements over the use of a single attention layer, but its reliance on simplistic aggregation assumptions often led to highlighting irrelevant tokens. LRP~\cite{bach2015pixel}, proposed to propagate gradients from the output layer to the beginning, considering all the components in the transformer's layers beyond the attention layers.
Chefer et al.~\cite{chefer2021transformer} presented Transformer Attribution (T-Attr), a class-specific Deep Taylor Decomposition method in which relevance propagation is applied for positive and negative attributions. More recently, the authors introduced Generic Attention Explainability (GAE) ~\cite{chefer2021generic}, a generalization of T-Attr for explaining Bi-Modal transformers. Both T-Attr and GAE are considered state-of-the-art methods for explaining ViT models and have been shown to outperform multiple strong baselines such as LRP, partial LRP~\cite{voita2019analyzing}, ViT-GC~\cite{chefer2021transformer}, and Rollout~\cite{abnar2020quantifying}.
IIA distinguishes itself from the aforementioned approaches in three key ways: Firstly, IIA introduces and utilizes the Gradient Rollout (GR) - a variant of Rollout that combines attention matrices with their gradients. Secondly, IIA employs GR as the integrand in its iterative integration process, conducting integration across interpolated attention matrices. Lastly, IIA stands out as a universal method, capable of generating explanations for both CNNs and ViTs.

\section{Iterated Integrated Attributions}
\label{sec:method}
We start by describing the problem setup. Then, we briefly overview IG~\cite{SundararajanTY17} and continue to describe \model in detail.
\subsection{Problem Setup}
Let $\x \in \R^{c_0 \times p_0 \times q_0}$ be an input image. We define a generic neural network model with $L$ intermediate layers, each is a function $h^l$ ($1\leq l \leq L$) that outputs $\x^l:=h^l(\x^{l-1})$, with $\x^0:=\x$. The final layer is a classification head $f$ that produces the prediction $f(\x^L)$, and the score for the class $y$ is given by $f_y(\x^L)$. Additionally, we define the application of the neural network to the input $\x$ by
\begin{equation}
\label{eq:phi}
\phi(\x)=f(\x^L).
\end{equation}
For example, if $\phi$ is a ResNet (ViT) model, each $h^l$ would be implemented as a residual convolutional (transformer encoder) block. Our goal is to generate an explanation map $\m\in\R^{p_0 \times q_0}$ that quantifies the attribution of each element in $\x$ to the prediction $\phi(\x)$. The attribution can be computed w.r.t. $\phi_y(\x)$ - the score assigned to the class $y$. Typically, the class of interest $y$ is either set to the target (ground-truth) class or to the predicted class, which is the class receiving the highest score in $\phi(\x)$.

\subsection{IG}
In what follows, we quickly overview IG~\cite{SundararajanTY17}, which is a special case of IIA. Given the input $\x$ and a reference $\r \in \R^{c_0 \times p_0 \times q_0}$ (that is designed to represent missing information, hence usually set to the zero image), we define a linear interpolant by 
\begin{equation}
\label{eq:v-ig}
\v=\r+a(\x-\r),
\end{equation}
with $a\in[0,1]$. IG produces an explanation map by integrating gradients along the linear path between $\r$ and $\x$ as follows:
\begin{equation}
\label{eq:ig}
\m_{IG}=\int_0^1 \frac{\partial \phi_y(\v)}{\partial \v} \circ \frac{\partial \v}{\partial a}da=(\x-\r) \circ \int_0^1 \frac{\partial \phi_y(\v)}{\partial \v}da,
\end{equation}

where $\circ$ stands for the Hadamard (elementwise) product. In practice, the integral in Eq.~\ref{eq:ig} is numerically approximated as follows:
\begin{equation}
\label{eq:ig-approx}
\m_{IG}\approx \frac{\x-\r}{n} \circ \sum_{k=1}^n \frac{\partial \phi_y(\v)}{\partial \v},
\end{equation}
by setting $a=\frac{k}{n}$ in Eq.~\ref{eq:v-ig}. The approximation in Eq.~\ref{eq:ig-approx} simply sums the gradients of $n$ interpolants on the linear path from $\r$ to $\x$. Finally, since $\m_{IG}$ is in $\R^{c_0 \times p_0 \times q_0}$ (typically, $c_0=3$ since $\x$ is a RGB image), mean reduction along the channel axis is performed to obtain a 2D explanation map.

\subsection{IIA - A Generic Formulation}
IIA diverges from IG in several aspects: First, IIA does not confine gradient computation to the input $\mathbf{x}$. In fact, recent studies have suggested that gradients derived from internal activation maps can yield improved explanation maps~\cite{selvaraju2017grad, chattopadhay2018grad,Fu2020AxiombasedGT}. Secondly, IIA employs an iterated integral across multiple intermediate representations (such as activation or attention maps) generated during the network's forward pass. This enables the iterative accumulation of gradients w.r.t. the representations of interest. Lastly, unlike IG, IIA does not restrict the integrand to plain gradients, but encompasses a function of the entire set of representations produced by the network and their gradients. In this section, we assume a generic neural network model. In Sec.~\ref{sec:impl}, we describe the utilization of IIA for CNN and ViT models. 

As outlined above, IIA utilizes linear interpolations on the intermediate representations generated during the forward propagation of the input through the layers of the model. In order to incorporate interpolation, we modify the computation in the $l$-th layer to accommodate an interpolation to the intermediate representation of interest (produced as part of the computational pipeline of $h^l$). To facilitate the formulation of the IIA approach, we introduce a set of notations: First, the input to the $l$-th layer undergoes processing by a function $u^l$ to obtain the intermediate representation of interest, denoted as $\u^l$. Subsequently, an interpolation step is (optionally) performed to derive the interpolant $\v^l$ (interpolated version of $\u^l$). Finally, the interpolant $\v^l$ is processed by a function $v^l$ that completes the original computational pipeline, yielding the input to the subsequent layer in the model. This entire process can be expressed mathematically using the following equations:
\begin{equation}
\label{eq:h}
\h^l=v^l(\v^l),
\end{equation}
with
\begin{equation}
\label{eq:v}
\v^l=\r^l + (a_l)^{b_l}(\u^l - \r^l),
\end{equation}
and
\begin{equation}
\label{eq:u}
\u^l=u^l(\h^{l-1}).
\end{equation}
The rationale behind Eqs.~\ref{eq:h}-\ref{eq:u} is as follows: $u^l$ is a function that computes the intermediate representation of interest $\u^l$ (the representation that is to be interpolated) based on the input to the $l$-th layer $\h^{l-1}$. $\u^l$ is further subtracted by a corresponding reference\footnote{The  reference should represent missing information. Other possible choices include (but not limited to) the null representation or random noise.} representation $\r^l=\texttt{min}(\u^l)$, which is the minimum value in each channel of $\u^l$ that is subsequently broadcast to a tensor with the same dimensions as $\u^l$. Additionally, in Eq.~\ref{eq:v}, $b_l$ is an indicator parameter that determines whether the interpolation is effectively applied to $\u^l$ during the propagation via the $l$-th layer in the model ($b_l=1$) or not ($b_l=0$), and $a_l\in[0,1]$ controls the interpolation step, hence playing a similar role as $a$ from Eq.~\ref{eq:ig}, resulting in the interpolant $\v^l$. Finally, $v^l$ is a function that receives the (interpolated) intermediate representation $\v^l$ and completes the required computation for producing the expected output from the $l$-th layer. Hence, the dimensions of $\h^l$ must match those of $h^l(\x^{l-1})$. Moreover, if $b_j=0$ for all $j\leq l$, $u^l=h^l$ and $v^l$ is the identity mapping, then $\h^l$ and $h^l(\x^{l-1})$ are identical.
Note that the implementation of $u^l$, $v^l$, and the choice of representations to be interpolated all vary based on the model's architecture (as will be detailed in Sec.~\ref{sec:impl}).

We further define $\b=[b_0,...,b_L]\in \{0,1\}^{L+1}$, $\h^{-1}=\x$, and set $u^0$ and $v^0$ to the identity mapping. Therefore, we have
\begin{equation}
\label{eq:u0-v0}
\u^0=\x \,\, \text{and} \,\, \h^0=\v^0.
\end{equation}
Finally, the IIA explanation map is defined as follows:
\begin{equation}
\label{eq:IIA-vanilla}
\m^l_\b=\int_0^1 \int_0^1 \dots (\u^l - \r^l) \circ \int_0^1 \q^l\, da_l \dots da_0,
\end{equation}
where the integrand $\q^l$ is a function of the first $l$ intermediate representations produced by the model (including the input representation) and their gradients. %

It is worth exploring the versatility of Eq.~\ref{eq:IIA-vanilla}: $\q^l$ determines the integrand that is a function of the participating representations and their gradients from the first $l$ layers in the model. $\b$ determines which of the representations produced by the first $l$ layers are effectively interpolated: if $b_j=1$ $(0\leq j \leq l)$, then the integration is effectively applied w.r.t. the variable $a_j$, otherwise $b_j=0$ and both Eqs.~\ref{eq:v} and \ref{eq:IIA-vanilla} become agnostic to $a_j$.
For example, one can observe that by setting $\b_0=[1,0...,0]$, $l=0$, $u^l=h^l$ (for $l>0$), $\q^l=\frac{\partial f_y(\h^L)}{\partial \v^l}$, and $v^l$ to the identity mapping, Eq.~\ref{eq:IIA-vanilla} (IIA) degenerates to Eq.~\ref{eq:ig} (IG) as follows: 
\begin{equation*}
\begin{split}
\label{eq:IIG-degen}
\m^0_{\b_0}&=(\u^0-\r^0) \circ \int_0^1 \frac{\partial f_y(\h^L)}{\partial \v^0}\,da_0\\&= (\x-\r^0) \circ \int_0^1 \frac{\partial \phi_y(\v^0)}{\partial \v^0}\,da_0.
\end{split}
\end{equation*}
The first equality follows from Eq.~\ref{eq:IIA-vanilla}, and the second is due to Eqs.~\ref{eq:phi} and \ref{eq:u0-v0}. Finally, by dropping the zero index, we receive Eq.~\ref{eq:ig}.

IIA (Eq.~\ref{eq:IIA-vanilla}) provides the freedom to run over multiple interpolated representations (including the input) in an iterative manner. Once $l$ is set, the integrand $\q^l$ changes based on the interpolated representations $\h^j$ ($j\leq l$) in the preceding layers that participate in the interpolation process, where participation is determined by the indicator vector $\b$. For example, if we set $l=L$ and $\b=[1,1,\ldots,1]$, $\m_b^L$ will be the outcome of a $L$ iterated integrals over $\q^L$. Thus, in computing $\m_b^L$, all the intermediate representations within the network (including the input) are iteratively interpolated. 

In practice, $\m_b^l$ is numerically approximated using:
\begin{equation}
\begin{split}
\label{eq:IIA-vanilla-approx}
\m^l_\b&\approx\frac{1}{n} \sum_{k_0=1}^n \frac{1}{n} \sum_{k_1=1}^n \dots \frac{1}{n} (\u^l-\r^l) \circ \sum_{k_l=1}^n  \q^l \\&= \frac{1}{n^\beta} \sum_{k_0=1}^{n^{b_0}} \sum_{k_1=1}^{n^{b_1}} \dots (\u^l-\r^l) \circ \sum_{k_l=1}^{n^{b_l}}  \q^l\,,
\end{split}
\end{equation}
$\beta=\sum_{i=0}^l b_i$, and $a_j=\frac{k_j}{n}$ (Eq.~\ref{eq:v}).
Again, Eq.~\ref{eq:IIA-vanilla-approx} degenerates to Eq.~\ref{eq:ig-approx} for $\b_0=[1,0...,0]$ and $l=0$.
Note that if $\q^l$ is not a 2D tensor, a subsequent mean reduction step is required to obtain a 2D explanation map (followed by a resize operation to align with the spatial dimensions of the input $\x$, if needed).

\subsection{IIA Implementation}
\label{sec:impl}
\paragraph{CNN Models} In CNNs, $\phi$ follows a CNN architecture (e.g., ResNet~\cite{he2016deep}). In this case, all $h^l$ are residual convolutional blocks producing 3D tensors, i.e., activation maps. In our implementation, we choose to apply the interpolation on the activation maps, hence we set all $v^l$ to the identity mapping, $u^l=h^l$, and the \texttt{min} reduction operation in the computation of $\r^l$ is applied channel-wise (followed by broadcasting). Additionally, we set the integrand $\q^l=\v^l \circ \frac{\partial f(\h^L)}{\partial \v^l}$. The motivation for this choice is as follows: $\v^l$ is the (interpolated) activation map that highlights regions where filters are activated, facilitating pattern detection. Its gradient quantifies the attribution level of the particular class of interest to each element in the activation map. Thus, we anticipate that areas where both the gradient and activation exhibit substantial magnitude with a consistent sign will yield effective explanations. This characteristic is achieved through the Hadamard product between $\v^l$ and its gradient. Finally, we apply a mean reduction to the channel axis, followed by a resize operation to obtain a 2D explanation map.
\paragraph{ViT Models} In the case of ViT~\cite{dosovitskiy2020image_vit}, the input $\x$ is a 2D tensor corresponding to a sequence of tokens (vectors), where the first token is the \textsc{[CLS]} token, and the rest represent patches from the input image. 
In our implementation, we opted to interpolate the attention matrices. To this end, we set $u^l$ to the attention function which involves the softmax operation on the scaled dot-product between the query and key representations across multiple attention heads. Assuming there are $p$ attention heads, for each head, we perform interpolation on the attention matrix. In this process, the reference $\r^l$ is assigned as the zero tensor since all entries in the attention matrices are positive due to the softmax operation. Accordingly, $v^l$ continues the self-attention computation by multiplying the interpolated attention matrices with the value representations for each head. This is followed by the necessary computational steps that generate a new set of token representations for the subsequent transformer encoder layer~\cite{dosovitskiy2020image_vit}. Finally, we propose setting the integrand $\q^l$ to the Gradient Rollout (GR) - a variant of the Attention Rollout (AR) method~\cite{abnar2020quantifying}. Similarly to AR, GR amalgamates information from the \textsc{[CLS]} attention across all attention heads in the model. However, with a notable distinction, each (interpolated) attention matrix is substituted by the Hadamard product of the attention matrix and its corresponding gradient. The exact implementation of GR is detailed in our git repository.
Given that the output of GR is already in the form of a 2D tensor, only a subsequent resize operation is necessary to achieve an explanation map that corresponds to the spatial dimensions of the input image. Finally, it is noteworthy that our experimentation indicates that replacing the matrix product operation with the matrix sum (as part of the GR computation) leads to comparable performance.

Due to the large combinatorial space ($2^L$ possible combinations for $\b$), and the fact we evaluate on large models, in this work, we consider double and triple integration in our complete experiments.

For double integration (\textbf{IIA2}), we set $l=L$, and $\b=[1,0,0,...,0,0,1]$, in Eq.~\ref{eq:IIA-vanilla-approx}, i.e., $b_0=1$ and $b_L=1$, and the rest $b_l=0$ ($1<l<L$). This means IIA effectively interpolates over the input image and the activation (attention) maps from the last layer in the CNN (ViT) model. Interpolating on the input image, enables us to examine various interpolations of the image and study the significance of pixel features along the integration path. Moreover, integrating on the last layer allows us to explore the importance of the aggregated information from the different layers of the network, as it combines all the network's features.

For triple integration (\textbf{IIA3}), we further interpolate on the penultimate layer $L-1$, i.e., $b_0=1$, $b_{L-1}=1$, $b_L=1$, and the rest $b_l=0$ ($1<l<L$). This is motivated by the fact that the penultimate layer captures more comprehensive objects and features, as it is closer to the classification head. By including a broader aggregation of features, it assists in predicting specific classes. In contrast, earlier layers primarily focus on detecting low-level features such as edges. 

Finally, for both IIA2 and IIA3, we set $n=10$ and $l=L$ in Eq.~\ref{eq:IIA-vanilla-approx}, i.e., 10 interpolation steps for each selected layer, and the integrand is computed w.r.t. the last layer.

\subsection{Computational Complexity}
\label{subsubsec:comp_complex}
The computational complexity of IIA is determined by the order of the iterated integral being computed. We utilized the approximation from Eq.~\ref{eq:IIA-vanilla-approx}, which is based on nested sums (each comprising $n$ terms). Each term necessitates the application of $\q^l$, whose computational complexity varies based on the specific implementation. For instance, in Sec.~\ref{sec:impl}, $\q^l$ combines both activation (attention) maps and their gradients, leading to computations involving both forward and backward passes. 
Therefore, if the computational complexity of $\q^l$ is $\mathcal{O}(Q)$, the overall computational complexity of Eq.~\ref{eq:IIA-vanilla-approx} is $\mathcal{O}(n^{\beta}Q)$. Yet, the complexity induced by $n^{\beta}$ can be significantly reduced through the utilization of batch processing via GPUs. For example, in IIA2 (iterated integration on the input and the last layer), performing $n$ interpolations on the input in a batch is straightforward. Next, we can extend this process to internal layers: creating batches for all interpolations of each activation map, concatenating these batches into a single batch, propagating it from the last layer to the prediction head, and computing gradients of $f$ w.r.t. the activation maps. Formally, the runtime complexity of IIA can be expressed as $R(IIA_M)=(\sum_{m=1}^{M} \frac{n^m}{B} c_{i_{m-1},i_{m}}) + \frac{n^{M}}{B} c_{i_M, K}$, where $c_{i,j}$ denotes the cost of propagating the data from layer $i$ to layer $j$ (or backpropagating from $j$ to $i$), $K$ denotes the index of the prediction layer $f$, $n$ indicates the number of interpolation steps, $B$ is the maximal batch size that can be accommodated by the GPU, and $M$ is the number of layers where interpolation is effectively applied (e.g., in IIA2, $M=2$). Note that the first and second terms in $R(IIA_M)$ are the costs of the forward and backward passes, respectively. Assuming a GPU with $B \geq n^{M}$, it follows that $\frac{n^m}{B}=\mathcal{O}(1)$ for all $1\leq m \leq M$, resulting in the cost of $IIA_M$ being bounded by a \emph{single} forward-backward pass. For example, for IIA2 and IIA3 with $n=10$, having $B=100$ and $B=1000$, respectively, is adequate to achieve $\frac{n^{M}}{B}=\mathcal{O}(1)$, which should be manageable with a high performance GPU. In these scenarios, the runtimes of GC, IG, and IIA are comparable. Theoretically, if $B \geq n^{M}$, IIA can become faster than IG, since in IG the gradients are backpropagated through the entire network back to the \emph{input}, while in IIA2 gradients are backpropagated to the layer $i_{M}$ (usually one of the \emph{penultimate} layers). Lastly, distributing IIA computations across multiple machines can yield further speed-up.

\section{Experimental Setup and Results}
\label{sec:experiments}

Our evaluation include five models: ViT-Base (\textbf{ViT-B}), ViT-Small (\textbf{ViT-S})~\cite{dosovitskiy2020image_vit}, ResNet101 (\textbf{RN})~\cite{He2016DeepRL}, DenseNet201 (\textbf{DN})~\cite{Huang2017DenselyCC}, and ConvNext-Base (\textbf{CN})~\cite{Liu2022ACF}. Preprocessing details and links to all models are provided in our GitHub repository.

\vspace{-3mm}
\paragraph{Evaluation Tasks and Metrics}
We present an extensive evaluation of both explanation and segmentation tasks. It is worth noting that having superior segmentation accuracy does not necessarily equate to having superior explanatory proficiency. Nevertheless, we conduct segmentation tests to ensure comprehensive comparison with previous works~\cite{chefer2021transformer,chefer2021generic,jiang2021layercam,wang2020score}. The explanation metrics include Area Under the Curves (AUCs) of Positive (\textbf{POS}) and Negative (\textbf{NEG}) perturbations tests~\cite{chefer2021transformer}, AUC of the Insertion (\textbf{INS}) and Deletion (\textbf{DEL}) tests~\cite{petsiuk2018rise}, AUC of the Softmax Information Curve (\textbf{SIC}) and Accuracy Information Curve (\textbf{AIC})~\cite{kapishnikov2019xrai}, Average Drop Percentage (\textbf{ADP}), and Percentage Increase in Confidence (\textbf{PIC})~\cite{chattopadhay2018grad}. For POS, DEL, and ADP the lower the better, while for NEG, INS, SIC, AIC, and PIC the higher the better. 
The segmentation metrics include Pixel Accuracy (\textbf{PA}), mean-intersection-over-union (\textbf{mIoU}), mean-average-precision (\textbf{mAP}), and the mean-F1 score (\textbf{mF1})~\cite{chefer2021transformer}. 
A detailed description of the metrics is provided in Appendix~\ref{sec:metric_desc}. Finally, in Appendix~\ref{sec:res_sanity}, we provide extensive evaluation on sanity tests~\cite{adebayo2018sanity} that further validate IIA as a machinery for generating faithful explanation maps.

\vspace{-3mm}
\paragraph{Datasets}
Explanation maps are produced for the ImageNet~\cite{imagenet} ILSVRC 2012 (\textbf{IN}) validation set, consisting of 50K images from 1000 classes. We follow the same setup from~\cite{chefer2021transformer}, where for each image, an explanation map is produced twice: (1) w.r.t. the ground-truth class (\textbf{Target}) and (2) w.r.t. the class predicted by the model (\textbf{Predicted}), i.e., the class that received the highest score. Accordingly, results are reported for both the predicted and target classes. 
Segmentation tests are conducted on three datasets:
(1) ImageNet-Segmentation~\cite{guillaumin2014imagenet} (\textbf{IN-Seg}): This is a subset of ImageNet validation set consisting of 4,276 images from 445 classes for which annotated segmentations are available. (2) Microsoft Common Objects in COntext 2017~\cite{lin2014microsoft} (\textbf{COCO}): This is a validation set that contains 5,000 annotated segmentation images from 80 different classes. Some images consist of multi-label annotations (multiple annotated objects). In our evaluation, all annotated objects in the image are considered as the ground-truth. (3) PASCAL Visual Object Classes  2012~\cite{Everingham2009ThePV} (\textbf{VOC}): A validation set that contains annotated segmentations for 1,449 images from 20 classes.
\vspace{-3mm}
\paragraph{Evaluated Methods and Hyperparameter Setting}
The following explanation methods for CNN models are evaluated as baselines: (1) Grad-CAM (\textbf{GC})~\cite{selvaraju2017grad}. (2) Grad-CAM++ (\textbf{GC++})~\cite{chattopadhay2018grad}. (3) FullGrad (\textbf{FG})~\cite{Srinivas2019FullGradientRF}. (4) Ablation-CAM (\textbf{AC})~\cite{Desai2020AblationCAMVE}. (5) Layer-CAM (\textbf{LC})~\cite{jiang2021layercam}. (6) LIFT-CAM (\textbf{LIFT})~\cite{Jung2021TowardsBE}, a state-of-the-art method that was shown to outperform other strong baselines like ScoreCAM~\cite{Wang2020ScoreCAMSV}. (7) Integrated Gradients (\textbf{IG})~\cite{SundararajanTY17}. (8) Guided IG (\textbf{GIG})~\cite{kapishnikov2021guided}. (9) Blur IG (\textbf{BIG})~\cite{xu2020attribution}. (10) X-Grad-CAM (\textbf{XGC})~\cite{Fu2020AxiombasedGT}. 
For ViT models, we considered the following two methods: (11) Transformer Attribution (\textbf{T-Attr})~\cite{chefer2021transformer}, a state-of-the-art method that was shown to outperform a variety of other strong baselines such as LRP~\cite{binder2016layer}, partial LRP~\cite{voita2019analyzing}, Raw-Attention~\cite{chefer2021generic}, GC~\cite{chefer2021generic} for transformers, and Rollout. (12) Generic Attention Explainability (\textbf{GAE})~\cite{chefer2021generic} - This is another state-of-the-art method that was shown to outperform T-Attr on several metrics. When applied, hyperparameters for all methods were configured according to the recommended configuration by the authors. (13) Our generic IIA method is evaluated on both CNN and ViT models. A detailed description of the baselines is provided in Appendix~\ref{sec:base_desc}.

\begin{table*}[t]
\begin{center}
  \begin{small}
   \scalebox{1.02}{
    \begin{tabular}
    {@{}lc@{}lc@{}lc@{}lc@{}lc@{}lc@{}lc@{}lc@{}}%
    \toprule
      & & & \multicolumn{1}{l}{GC} & \multicolumn{1}{l}{GC++} & \multicolumn{1}{l}{LIFT} & \multicolumn{1}{l}{AC} & \multicolumn{1}{l}{IG}& \multicolumn{1}{l}{GIG}& \multicolumn{1}{l}{BIG} & \multicolumn{1}{l}{FG} & \multicolumn{1}{l}{LC} & \multicolumn{1}{l}{XGC} & \multicolumn{1}{l}{IIA2} & \multicolumn{1}{l}{IIA3}\\
    \midrule
    \multirow{16}{*}{RN}
    & \multirow{2}{*}{NEG} & \multicolumn{1}{l}{Predicted} & \multicolumn{1}{l}{\underline{56.41}} & \multicolumn{1}{l}{55.20} & \multicolumn{1}{l}{55.39} & \multicolumn{1}{l}{54.98} & \multicolumn{1}{l}{45.66} &\multicolumn{1}{l}{43.97} & \multicolumn{1}{l}{42.25}& \multicolumn{1}{l}{54.81} & \multicolumn{1}{l}{53.52} & \multicolumn{1}{l}{53.46} & \multicolumn{1}{l}{56.29} & \multicolumn{1}{l}{\textbf{56.63}} \\

         & & \multicolumn{1}{l}{Target} & \multicolumn{1}{l}{\underline{56.54}} & \multicolumn{1}{l}{55.95} & \multicolumn{1}{l}{55.23} & \multicolumn{1}{l}{55.46} & \multicolumn{1}{l}{42.02} &\multicolumn{1}{l}{41.93} & \multicolumn{1}{l}{41.22}& \multicolumn{1}{l}{54.65} & \multicolumn{1}{l}{54.19} & \multicolumn{1}{l}{53.93} & \multicolumn{1}{l}{56.22} & \multicolumn{1}{l}{\textbf{56.89}}\\

    & \multirow{2}{*}{POS} & \multicolumn{1}{l}{Predicted} & \multicolumn{1}{l}{17.82} & \multicolumn{1}{l}{18.01} & \multicolumn{1}{l}{17.53} & \multicolumn{1}{l}{19.38} & \multicolumn{1}{l}{17.24} &\multicolumn{1}{l}{17.68} & \multicolumn{1}{l}{17.44} & \multicolumn{1}{l}{18.06} & \multicolumn{1}{l}{17.92} & \multicolumn{1}{l}{21.02} & \multicolumn{1}{l}{\underline{16.62}} & \multicolumn{1}{l}{\textbf{16.19}}\\

         & & \multicolumn{1}{l}{Target} & \multicolumn{1}{l}{17.65} & \multicolumn{1}{l}{18.12}  & \multicolumn{1}{l}{17.48} & \multicolumn{1}{l}{19.73} & \multicolumn{1}{l}{16.93}&\multicolumn{1}{l}{17.48} & \multicolumn{1}{l}{17.26} & \multicolumn{1}{l}{17.89} &  \multicolumn{1}{l}{17.79} & \multicolumn{1}{l}{20.61} & \multicolumn{1}{l}{\underline{16.69}} & \multicolumn{1}{l}{\textbf{15.78}}\\

    & \multirow{2}{*}{INS} & \multicolumn{1}{l}{Predicted} & \multicolumn{1}{l}{\underline{48.14}} & \multicolumn{1}{l}{47.56} & \multicolumn{1}{l}{45.39} & \multicolumn{1}{l}{47.85} & \multicolumn{1}{l}{39.87}  &\multicolumn{1}{l}{37.92} & \multicolumn{1}{l}{36.04}& \multicolumn{1}{l}{42.68} & \multicolumn{1}{l}{46.11} & \multicolumn{1}{l}{43.26} & \multicolumn{1}{l}{48.01} & \multicolumn{1}{l}{\textbf{49.12}}\\

         & & \multicolumn{1}{l}{Target} & \multicolumn{1}{l}{\underline{48.22}} & \multicolumn{1}{l}{47.27} & \multicolumn{1}{l}{44.94} & \multicolumn{1}{l}{47.75} & \multicolumn{1}{l}{37.55} &\multicolumn{1}{l}{34.41} & \multicolumn{1}{l}{34.68}&  \multicolumn{1}{l}{43.08} & \multicolumn{1}{l}{45.91} & \multicolumn{1}{l}{43.13} & \multicolumn{1}{l}{48.05} & \multicolumn{1}{l}{\textbf{48.53}}\\

    & \multirow{2}{*}{DEL} & \multicolumn{1}{l}{Predicted} & \multicolumn{1}{l}{13.97} & \multicolumn{1}{l}{14.17} & \multicolumn{1}{l}{15.32} & \multicolumn{1}{l}{14.23} & \multicolumn{1}{l}{13.49} &\multicolumn{1}{l}{14.18} & \multicolumn{1}{l}{13.95}& \multicolumn{1}{l}{14.64} & \multicolumn{1}{l}{14.31} & \multicolumn{1}{l}{14.98} & \multicolumn{1}{l}{\underline{13.18}} & \multicolumn{1}{l}{\textbf{12.74}}\\

         & & \multicolumn{1}{l}{Target} & \multicolumn{1}{l}{13.63} & \multicolumn{1}{l}{13.94} & \multicolumn{1}{l}{15.48} & \multicolumn{1}{l}{15.05} & \multicolumn{1}{l}{13.46} &\multicolumn{1}{l}{14.31} & \multicolumn{1}{l}{14.26}& \multicolumn{1}{l}{14.98} & \multicolumn{1}{l}{13.71} & \multicolumn{1}{l}{14.72} & \multicolumn{1}{l}{\underline{12.82}} & \multicolumn{1}{l}{\textbf{12.16}}\\

     & \multirow{2}{*}{ADP} & \multicolumn{1}{l}{Predicted} & \multicolumn{1}{l}{17.87} & \multicolumn{1}{l}{16.91} & \multicolumn{1}{l}{18.03} & \multicolumn{1}{l}{16.19} & \multicolumn{1}{l}{37.52} &\multicolumn{1}{l}{35.28} & \multicolumn{1}{l}{40.85}& \multicolumn{1}{l}{21.06} & \multicolumn{1}{l}{24.34} & \multicolumn{1}{l}{17.02} & \multicolumn{1}{l}{\textbf{12.79}} & \multicolumn{1}{l}{\underline{12.84}}\\

         & & \multicolumn{1}{l}{Target} & \multicolumn{1}{l}{17.83} & \multicolumn{1}{l}{15.97} & \multicolumn{1}{l}{17.36} & \multicolumn{1}{l}{15.30} & \multicolumn{1}{l}{36.51} &\multicolumn{1}{l}{36.00} & \multicolumn{1}{l}{41.98}& \multicolumn{1}{l}{20.29} & \multicolumn{1}{l}{23.78} & \multicolumn{1}{l}{16.39} & \multicolumn{1}{l}{\textbf{12.31}} & \multicolumn{1}{l}{\underline{12.40}}\\

    & \multirow{2}{*}{PIC} & \multicolumn{1}{l}
    {Predicted} &  \multicolumn{1}{l}{36.69} & \multicolumn{1}{l}{36.53} & \multicolumn{1}{l}{35.95} & \multicolumn{1}{l}{35.52} & \multicolumn{1}{l}{19.94} &\multicolumn{1}{l}{18.72} & \multicolumn{1}{l}{24.53}& \multicolumn{1}{l}{31.59} & \multicolumn{1}{l}{35.43} & \multicolumn{1}{l}{36.18} & \multicolumn{1}{l}{\textbf{42.96}} & \multicolumn{1}{l}{\underline{42.91}}\\

         & & \multicolumn{1}{l}{Target} & \multicolumn{1}{l}{37.84} & \multicolumn{1}{l}{38.37} & \multicolumn{1}{l}{37.64} & \multicolumn{1}{l}{37.31} & \multicolumn{1}{l}{21.43} &\multicolumn{1}{l}{15.81} & \multicolumn{1}{l}{23.94}& \multicolumn{1}{l}{33.18} & \multicolumn{1}{l}{35.64} & \multicolumn{1}{l}{37.54} & \multicolumn{1}{l}{\textbf{45.21}} & \multicolumn{1}{l}{\underline{45.06}}\\

    & \multirow{2}{*}{SIC} & \multicolumn{1}{l}{Predicted} &  \multicolumn{1}{l}{76.91} & \multicolumn{1}{l}{76.44} & \multicolumn{1}{l}{76.73} & \multicolumn{1}{l}{73.36} & \multicolumn{1}{l}{54.67} & \multicolumn{1}{l}{55.04} & \multicolumn{1}{l}{56.98} & \multicolumn{1}{l}{75.35} & \multicolumn{1}{l}{73.93} & \multicolumn{1}{l}{72.64} & \multicolumn{1}{l}{\underline{78.52}} & \multicolumn{1}{l}{\textbf{79.92}}\\

             & & \multicolumn{1}{l}{Target} &  \multicolumn{1}{l}{76.87} & \multicolumn{1}{l}{76.62} & \multicolumn{1}{l}{76.81} & \multicolumn{1}{l}{73.55} & \multicolumn{1}{l}{51.54} & \multicolumn{1}{l}{54.87} & \multicolumn{1}{l}{55.23} & \multicolumn{1}{l}{75.39} & \multicolumn{1}{l}{73.71} & \multicolumn{1}{l}{72.71} & \multicolumn{1}{l}{\underline{78.13}} & \multicolumn{1}{l}{\textbf{79.94}}\\

        & \multirow{2}{*}{AIC} & \multicolumn{1}{l}{Predicted} &  \multicolumn{1}{l}{74.36} & \multicolumn{1}{l}{71.97} & \multicolumn{1}{l}{72.76} & \multicolumn{1}{l}{70.35} & \multicolumn{1}{l}{51.92} & \multicolumn{1}{l}{53.38} & \multicolumn{1}{l}{53.36} & \multicolumn{1}{l}{71.49} & \multicolumn{1}{l}{65.77} & \multicolumn{1}{l}{69.85} & \multicolumn{1}{l}{\underline{75.49}} & \multicolumn{1}{l}{\textbf{76.12}}\\

             & & \multicolumn{1}{l}{Target} &  \multicolumn{1}{l}{72.49} & \multicolumn{1}{l}{71.42} & \multicolumn{1}{l}{73.45} & \multicolumn{1}{l}{70.48} & \multicolumn{1}{l}{52.71} & \multicolumn{1}{l}{52.54} & \multicolumn{1}{l}{54.24} & \multicolumn{1}{l}{71.38} & \multicolumn{1}{l}{66.18} & \multicolumn{1}{l}{70.24} & \multicolumn{1}{l}{\underline{75.88}} & \multicolumn{1}{l}{\textbf{76.59}}\\
        \bottomrule
    \midrule
    \multirow{16}{*}{CN}

    & \multirow{2}{*}{NEG} & \multicolumn{1}{l}{Predicted} & \multicolumn{1}{l}{52.86} & \multicolumn{1}{l}{53.82} & \multicolumn{1}{l}{53.98} & \multicolumn{1}{l}{53.68} & \multicolumn{1}{l}{45.24} &\multicolumn{1}{l}{41.43} & \multicolumn{1}{l}{40.72}& \multicolumn{1}{l}{52.06} & \multicolumn{1}{l}{54.12} & \multicolumn{1}{l}{52.13} & \multicolumn{1}{l}{\underline{55.94}} & \multicolumn{1}{l}{\textbf{57.19}}\\

         & & \multicolumn{1}{l}{Target} & \multicolumn{1}{l}{53.02} & \multicolumn{1}{l}{53.05} & \multicolumn{1}{l}{53.24} & \multicolumn{1}{l}{53.27} & \multicolumn{1}{l}{44.56} &\multicolumn{1}{l}{42.12} & \multicolumn{1}{l}{40.03}& \multicolumn{1}{l}{52.65} & \multicolumn{1}{l}{53.21} & \multicolumn{1}{l}{52.91} & \multicolumn{1}{l}{\underline{56.61}} & \multicolumn{1}{l}{\textbf{57.34}}\\
    
    & \multirow{2}{*}{POS} & \multicolumn{1}{l}{Predicted} &  \multicolumn{1}{l}{17.52} & \multicolumn{1}{l}{17.85} & \multicolumn{1}{l}{18.23} & \multicolumn{1}{l}{18.19} & \multicolumn{1}{l}{17.42} &\multicolumn{1}{l}{18.03} & \multicolumn{1}{l}{18.14}& \multicolumn{1}{l}{18.26} & \multicolumn{1}{l}{17.58} & \multicolumn{1}{l}{20.83} & \multicolumn{1}{l}{\underline{15.67}} & \multicolumn{1}{l}{\textbf{15.28}}\\

         & & \multicolumn{1}{l}{Target} & \multicolumn{1}{l}{17.34} & \multicolumn{1}{l}{17.51} & \multicolumn{1}{l}{18.05} & \multicolumn{1}{l}{18.41} & \multicolumn{1}{l}{17.53}
         &\multicolumn{1}{l}{17.32} & \multicolumn{1}{l}{17.61}& \multicolumn{1}{l}{17.92} & \multicolumn{1}{l}{18.03} & \multicolumn{1}{l}{18.12} & \multicolumn{1}{l}{\underline{15.25}} & \multicolumn{1}{l}{\textbf{15.21}}\\

    & \multirow{2}{*}{INS} & \multicolumn{1}{l}{Predicted} &  \multicolumn{1}{l}{45.65} & \multicolumn{1}{l}{45.19} & \multicolumn{1}{l}{43.86} & \multicolumn{1}{l}{49.18} & \multicolumn{1}{l}{37.22} 
    &\multicolumn{1}{l}{32.99} & \multicolumn{1}{l}{31.02}& \multicolumn{1}{l}{42.01} & \multicolumn{1}{l}{44.14} & \multicolumn{1}{l}{42.07} & \multicolumn{1}{l}{\underline{50.36}} & \multicolumn{1}{l}{\textbf{51.23}}\\

         & & \multicolumn{1}{l}{Target} & \multicolumn{1}{l}{46.21} & \multicolumn{1}{l}{45.27} & \multicolumn{1}{l}{43.94} & \multicolumn{1}{l}{49.75} & \multicolumn{1}{l}{36.83} 
         &\multicolumn{1}{l}{33.58} & \multicolumn{1}{l}{33.92}& \multicolumn{1}{l}{42.08} & \multicolumn{1}{l}{44.91} & \multicolumn{1}{l}{42.14} & \multicolumn{1}{l}{\underline{50.91}} & \multicolumn{1}{l}{\textbf{51.45}}\\

         & \multirow{2}{*} {DEL} & 
         \multicolumn{1}{l}{Predicted} & \multicolumn{1}{l}{13.43} & \multicolumn{1}{l}{14.17} & \multicolumn{1}{l}{15.18} & \multicolumn{1}{l}{14.73} & \multicolumn{1}{l}{12.36} 
         &\multicolumn{1}{l}{13.08} & \multicolumn{1}{l}{13.29}& \multicolumn{1}{l}{14.21} & \multicolumn{1}{l}{13.64} & \multicolumn{1}{l}{14.78} & \multicolumn{1}{l}{\underline{11.68}} & \multicolumn{1}{l}{\textbf{11.29}}\\

             & & \multicolumn{1}{l}{Target} & \multicolumn{1}{l}{13.32} & \multicolumn{1}{l}{14.39} & \multicolumn{1}{l}{14.86} & \multicolumn{1}{l}{14.44} & \multicolumn{1}{l}{12.83}
             &\multicolumn{1}{l}{13.45} & \multicolumn{1}{l}{13.69}& \multicolumn{1}{l}{14.55} & \multicolumn{1}{l}{14.28} & \multicolumn{1}{l}{14.29} & \multicolumn{1}{l}{\underline{11.24}} & \multicolumn{1}{l}{\textbf{10.80}}\\

         & \multirow{2}{*}{ADP} & \multicolumn{1}{l}{Predicted} & \multicolumn{1}{l}{22.46} & \multicolumn{1}{l}{22.35} & \multicolumn{1}{l}{29.13} & \multicolumn{1}{l}{24.38} & \multicolumn{1}{l}{36.98} 
         &\multicolumn{1}{l}{35.79} & \multicolumn{1}{l}{41.73}& \multicolumn{1}{l}{30.75} & \multicolumn{1}{l}{37.62} & \multicolumn{1}{l}{25.68} & \multicolumn{1}{l}{\underline{16.73}} & \multicolumn{1}{l}{\textbf{16.47}}\\

             & & \multicolumn{1}{l}{Target} &  \multicolumn{1}{l}{22.39} & \multicolumn{1}{l}{21.13} & \multicolumn{1}{l}{28.06} & \multicolumn{1}{l}{23.03} & \multicolumn{1}{l}{35.62}
             &\multicolumn{1}{l}{34.12} & \multicolumn{1}{l}{40.82}& \multicolumn{1}{l}{29.64} & \multicolumn{1}{l}{36.61} & \multicolumn{1}{l}{24.74} & \multicolumn{1}{l}{\underline{16.28}} & \multicolumn{1}{l}{\textbf{15.94}}\\

        & \multirow{2}{*}{PIC} & \multicolumn{1}{l}{Predicted} &    \multicolumn{1}{l}{23.16} & \multicolumn{1}{l}{24.42} & \multicolumn{1}{l}{22.34} & \multicolumn{1}{l}{24.59} & \multicolumn{1}{l}{17.65} 
        &\multicolumn{1}{l}{13.12} & \multicolumn{1}{l}{20.69}& \multicolumn{1}{l}{22.13} & \multicolumn{1}{l}{22.17} & \multicolumn{1}{l}{23.26} & \multicolumn{1}{l}{\underline{27.11}} & \multicolumn{1}{l}{\textbf{27.44}}\\

             & & \multicolumn{1}{l}{Target} & \multicolumn{1}{l}{24.53} & \multicolumn{1}{l}{24.26} & \multicolumn{1}{l}{22.59} & \multicolumn{1}{l}{24.33} & \multicolumn{1}{l}{18.15} 
             &\multicolumn{1}{l}{13.46} & \multicolumn{1}{l}{20.48}& \multicolumn{1}{l}{23.93} & \multicolumn{1}{l}{22.38} & \multicolumn{1}{l}{23.59} & \multicolumn{1}{l}{\underline{27.95}} & \multicolumn{1}{l}{\textbf{28.13}}\\

        & \multirow{2}{*}{SIC} & \multicolumn{1}{l}{Predicted} &  \multicolumn{1}{l}{65.93} & \multicolumn{1}{l}{67.94} & \multicolumn{1}{l}{54.75} & \multicolumn{1}{l}{63.95} & \multicolumn{1}{l}{53.36} & \multicolumn{1}{l}{58.35} & \multicolumn{1}{l}{57.27} & \multicolumn{1}{l}{62.84} & \multicolumn{1}{l}{69.11} & \multicolumn{1}{l}{59.12} & \multicolumn{1}{l}{\underline{69.63}} & \multicolumn{1}{l}{\textbf{70.46}}\\

             & & \multicolumn{1}{l}{Target} &  \multicolumn{1}{l}{66.86} & \multicolumn{1}{l}{67.63} & \multicolumn{1}{l}{56.22} & \multicolumn{1}{l}{64.78} & \multicolumn{1}{l}{53.48} & \multicolumn{1}{l}{58.48} & \multicolumn{1}{l}{57.40} & \multicolumn{1}{l}{63.93} & \multicolumn{1}{l}{68.93} & \multicolumn{1}{l}{59.09} & \multicolumn{1}{l}{\underline{69.43}} & \multicolumn{1}{l}{\textbf{70.21}}\\

        & \multirow{2}{*}{AIC} & \multicolumn{1}{l}{Predicted} &  \multicolumn{1}{l}{75.64} & \multicolumn{1}{l}{75.52} & \multicolumn{1}{l}{57.06} & \multicolumn{1}{l}{71.53} & \multicolumn{1}{l}{51.68} & \multicolumn{1}{l}{55.82} & \multicolumn{1}{l}{53.82} & \multicolumn{1}{l}{67.15} & \multicolumn{1}{l}{75.41} & \multicolumn{1}{l}{62.38} & \multicolumn{1}{l}{\underline{77.89}} & \multicolumn{1}{l}{\textbf{78.75}}\\

        & & \multicolumn{1}{l}{Target} &  \multicolumn{1}{l}{76.92} & \multicolumn{1}{l}{75.81} & \multicolumn{1}{l}{60.82} & \multicolumn{1}{l}{71.85} & \multicolumn{1}{l}{50.98} & \multicolumn{1}{l}{55.25} & \multicolumn{1}{l}{53.86} & \multicolumn{1}{l}{69.53} & \multicolumn{1}{l}{74.12} & \multicolumn{1}{l}{61.78} & \multicolumn{1}{l}{\underline{77.62}} & \multicolumn{1}{l}{\textbf{78.64}}\\
        \bottomrule
\midrule
    \multirow{16}{*}{DN}

    & \multirow{2}{*}{NEG} & \multicolumn{1}{l}{Predicted} & \multicolumn{1}{l}{\underline{57.40}} & \multicolumn{1}{l}{57.16} & \multicolumn{1}{l}{58.01} & \multicolumn{1}{l}{56.63} & \multicolumn{1}{l}{40.74} 
    &\multicolumn{1}{l}{37.31} & \multicolumn{1}{l}{36.67}& \multicolumn{1}{l}{56.79} & \multicolumn{1}{l}{56.96} & \multicolumn{1}{l}{55.74} & \multicolumn{1}{l}{57.32} & \multicolumn{1}{l}{\textbf{58.01}}\\
  
         & & \multicolumn{1}{l}{Target} & 
         \multicolumn{1}{l}{\underline{58.56}} & \multicolumn{1}{l}{58.33} & \multicolumn{1}{l}{58.07} & \multicolumn{1}{l}{57.78} & \multicolumn{1}{l}{41.32} 
         &\multicolumn{1}{l}{41.95} & \multicolumn{1}{l}{40.88}& \multicolumn{1}{l}{58.05} & \multicolumn{1}{l}{58.49} & \multicolumn{1}{l}{56.87} & \multicolumn{1}{l}{58.51} & \multicolumn{1}{l}{\textbf{59.15}}\\

    & \multirow{2}{*}{POS} & \multicolumn{1}{l}{Predicted} &  \multicolumn{1}{l}{17.75} & \multicolumn{1}{l}{17.81} & \multicolumn{1}{l}{18.87} & \multicolumn{1}{l}{18.67} & \multicolumn{1}{l}{17.31} 
    &\multicolumn{1}{l}{17.46} & \multicolumn{1}{l}{17.38}& \multicolumn{1}{l}{17.84} & \multicolumn{1}{l}{17.62} & \multicolumn{1}{l}{18.67} & \multicolumn{1}{l}{\underline{16.82}} & \multicolumn{1}{l}{\textbf{16.51}} \\

         & & \multicolumn{1}{l}{Target} & \multicolumn{1}{l}{17.52} & \multicolumn{1}{l}{17.78} & \multicolumn{1}{l}{17.79} & \multicolumn{1}{l}{19.69} & \multicolumn{1}{l}{17.00}
         &\multicolumn{1}{l}{17.41} & \multicolumn{1}{l}{17.34}& \multicolumn{1}{l}{17.46} & \multicolumn{1}{l}{16.92} & \multicolumn{1}{l}{18.57} & \multicolumn{1}{l}{\underline{16.63}} & \multicolumn{1}{l}{\textbf{16.01}}\\

    & \multirow{2}{*}{INS} & \multicolumn{1}{l}{Predicted} & \multicolumn{1}{l}{\underline{51.09}} & \multicolumn{1}{l}{50.89} & \multicolumn{1}{l}{50.63} & \multicolumn{1}{l}{50.41} & \multicolumn{1}{l}{37.58} 
    &\multicolumn{1}{l}{33.31} & \multicolumn{1}{l}{31.32}& \multicolumn{1}{l}{50.44} & \multicolumn{1}{l}{50.60} & \multicolumn{1}{l}{49.62} & \multicolumn{1}{l}{50.98} & \multicolumn{1}{l}{\textbf{51.86}}\\

         & & \multicolumn{1}{l}{Target} & \multicolumn{1}{l}{\underline{51.98}} & \multicolumn{1}{l}{51.64} & \multicolumn{1}{l}{50.31} & \multicolumn{1}{l}{50.56} & \multicolumn{1}{l}{38.94}
         &\multicolumn{1}{l}{34.11} & \multicolumn{1}{l}{32.76}& \multicolumn{1}{l}{51.61} & \multicolumn{1}{l}{49.76} & \multicolumn{1}{l}{49.28} & \multicolumn{1}{l}{51.90} & \multicolumn{1}{l}{\textbf{52.65}}\\

         & \multirow{2}{*}{DEL} & \multicolumn{1}{l}{Predicted} & \multicolumn{1}{l}{13.61} & \multicolumn{1}{l}{13.63} & \multicolumn{1}{l}{13.29} & \multicolumn{1}{l}{15.31} & \multicolumn{1}{l}{13.26} 
         &\multicolumn{1}{l}{13.27} & \multicolumn{1}{l}{13.54}& \multicolumn{1}{l}{14.34} & \multicolumn{1}{l}{13.85} & \multicolumn{1}{l}{14.75} & \multicolumn{1}{l}{\underline{13.02}} & \multicolumn{1}{l}{\textbf{12.79}}\\

             & & \multicolumn{1}{l}{Target} & \multicolumn{1}{l}{13.42} & \multicolumn{1}{l}{13.57} & \multicolumn{1}{l}{13.36} & \multicolumn{1}{l}{15.21} & \multicolumn{1}{l}{13.12}
             &\multicolumn{1}{l}{13.84} & \multicolumn{1}{l}{13.68}& \multicolumn{1}{l}{14.18} & \multicolumn{1}{l}{13.69} & \multicolumn{1}{l}{14.37} & \multicolumn{1}{l}{\underline{12.19}} & \multicolumn{1}{l}{\textbf{11.93}}\\

         & \multirow{2}{*}{ADP} & \multicolumn{1}{l}{Predicted} & \multicolumn{1}{l}{17.46} & \multicolumn{1}{l}{17.01} & \multicolumn{1}{l}{19.45} & \multicolumn{1}{l}{17.13} & \multicolumn{1}{l}{35.61}
         &\multicolumn{1}{l}{34.51} & \multicolumn{1}{l}{40.04}& \multicolumn{1}{l}{20.21} & \multicolumn{1}{l}{24.23} & \multicolumn{1}{l}{19.59} & \multicolumn{1}{l}{\textbf{13.42}} & \multicolumn{1}{l}{\underline{13.56}}\\

             & & \multicolumn{1}{l}{Target} &  \multicolumn{1}{l}{17.52} & \multicolumn{1}{l}{16.06} & \multicolumn{1}{l}{18.76} & \multicolumn{1}{l}{16.21} & \multicolumn{1}{l}{29.72} 
             &\multicolumn{1}{l}{29.14} & \multicolumn{1}{l}{34.74}& \multicolumn{1}{l}{19.35} & \multicolumn{1}{l}{23.59} & \multicolumn{1}{l}{18.88} & \multicolumn{1}{l}{\underline{13.95}} & \multicolumn{1}{l}{\textbf{13.93}}\\

        & \multirow{2}{*}{PIC} & \multicolumn{1}{l}{Predicted} &  \multicolumn{1}{l}{34.68} & \multicolumn{1}{l}{35.21} & \multicolumn{1}{l}{34.13} & \multicolumn{1}{l}{31.22} & \multicolumn{1}{l}{22.35} 
        &\multicolumn{1}{l}{16.62} & \multicolumn{1}{l}{26.18}& \multicolumn{1}{l}{31.05} & \multicolumn{1}{l}{33.81} & \multicolumn{1}{l}{30.39} & \multicolumn{1}{l}{\underline{39.54}} & \multicolumn{1}{l}{\textbf{39.69}}\\

             & & \multicolumn{1}{l}{Target} & \multicolumn{1}{l}{34.82} & \multicolumn{1}{l}{35.38} & \multicolumn{1}{l}{34.59} & \multicolumn{1}{l}{31.85} & \multicolumn{1}{l}{23.96} 
             &\multicolumn{1}{l}{20.56} & \multicolumn{1}{l}{23.51}& \multicolumn{1}{l}{31.33} & \multicolumn{1}{l}{33.95} & \multicolumn{1}{l}{31.32} & \multicolumn{1}{l}{\textbf{39.98}} & \multicolumn{1}{l}{\underline{39.83}}\\

        & \multirow{2}{*}{SIC} & \multicolumn{1}{l}{Predicted} &  \multicolumn{1}{l}{75.62} & \multicolumn{1}{l}{74.75} & \multicolumn{1}{l}{74.72} & \multicolumn{1}{l}{73.94} & \multicolumn{1}{l}{54.59} & \multicolumn{1}{l}{58.55} & \multicolumn{1}{l}{57.66} & \multicolumn{1}{l}{72.93} & \multicolumn{1}{l}{74.34} & \multicolumn{1}{l}{73.94} & \multicolumn{1}{l}{\underline{77.71}} & \multicolumn{1}{l}{\textbf{78.13}}\\

             & & \multicolumn{1}{l}{Target} &  \multicolumn{1}{l}{75.79} & \multicolumn{1}{l}{74.91} & \multicolumn{1}{l}{74.35} & \multicolumn{1}{l}{73.31} & \multicolumn{1}{l}{53.45} & \multicolumn{1}{l}{59.02} & \multicolumn{1}{l}{56.85} & \multicolumn{1}{l}{73.64} & \multicolumn{1}{l}{73.93} & \multicolumn{1}{l}{74.22} & \multicolumn{1}{l}{\underline{76.85}} & \multicolumn{1}{l}{\textbf{77.27}}\\

        & \multirow{2}{*}{AIC} & \multicolumn{1}{l}{Predicted} &  \multicolumn{1}{l}{74.22} & \multicolumn{1}{l}{71.82} & \multicolumn{1}{l}{72.65} & \multicolumn{1}{l}{70.21} & \multicolumn{1}{l}{54.74} & \multicolumn{1}{l}{54.56} & \multicolumn{1}{l}{56.08} & \multicolumn{1}{l}{70.63} & \multicolumn{1}{l}{71.82} & \multicolumn{1}{l}{70.12} & \multicolumn{1}{l}{\underline{75.22}} & \multicolumn{1}{l}{\textbf{77.16}}\\

             & & \multicolumn{1}{l}{Target} &  \multicolumn{1}{l}{74.18} & \multicolumn{1}{l}{72.14} & \multicolumn{1}{l}{73.29} & \multicolumn{1}{l}{70.97} & \multicolumn{1}{l}{54.91} & \multicolumn{1}{l}{54.77} & \multicolumn{1}{l}{56.25} & \multicolumn{1}{l}{71.31} & \multicolumn{1}{l}{71.88} & \multicolumn{1}{l}{70.36} & \multicolumn{1}{l}{\underline{75.49}} & \multicolumn{1}{l}{\textbf{76.99}}\\
        \bottomrule
      \end{tabular}}
  \end{small}
  \end{center}
  \caption{Explanation tests results on the IN dataset (CNN models): For POS, DEL and ADP, lower is better. For NEG, INS, PIC, SIC and AIC, higher is better. See Sec.~\ref{sec:experiments} for details.}
  \label{tab:cnn_backbones_metrics_exp}
    \end{table*}
\begin{table}
  
  \begin{center}
  \begin{small}
  \scalebox{1}{
    \begin{tabular}{@{}lc@{}lc@{}lc@{}lc@{}}%
    \toprule
      & & & \multicolumn{1}{l}{T-Attr} & \multicolumn{1}{l}{GAE} & \multicolumn{1}{l}{IIA2} & \multicolumn{1}{l}{IIA3} \\
    \midrule
    \multirow{16}{*}{ViT-B}
    
        & \multirow{2}{*}{NEG} & \multicolumn{1}{l}
        {Predicted} & \multicolumn{1}{l}{{54.16}} & \multicolumn{1}{l}{{54.61}} & \multicolumn{1}{l}{\underline{56.01}} & \multicolumn{1}{l}{\textbf{57.68}}\\
             & & \multicolumn{1}{l}
             {Target} &  \multicolumn{1}{l}{55.04} & \multicolumn{1}{l}{55.67} & \multicolumn{1}{l}{\underline{57.47}} & \multicolumn{1}{l}{\textbf{58.31}}\\
        
        & \multirow{2}{*}{POS} & \multicolumn{1}{l}
        {Predicted} &  \multicolumn{1}{l}{17.03} & \multicolumn{1}{l}{17.32} & \multicolumn{1}{l}{\underline{15.19}} & \multicolumn{1}{l}{\textbf{14.96}}\\
             & & \multicolumn{1}{l}
             {Target} & \multicolumn{1}{l}{16.04}  &  \multicolumn{1}{l}{16.72}  &  \multicolumn{1}{l}{\underline{15.81}} & \multicolumn{1}{l}{\textbf{15.02}}\\

        & \multirow{2}{*}{INS} & \multicolumn{1}{l}
        {Predicted} &  \multicolumn{1}{l}{48.58} & \multicolumn{1}{l}{48.96} & \multicolumn{1}{l}{\underline{49.31}} & \multicolumn{1}{l}{\textbf{50.71}}\\
             & & \multicolumn{1}{l}
             {Target} &  \multicolumn{1}{l}{49.19}  &  \multicolumn{1}{l}{49.65}  & \multicolumn{1}{l}{\underline{50.49}} & \multicolumn{1}{l}{\textbf{51.26}}\\
        
        & \multirow{2}{*}{DEL} & \multicolumn{1}{l}
        {Predicted} & \multicolumn{1}{l}{{14.20}} & \multicolumn{1}{l}{{14.37}} & \multicolumn{1}{l}{\underline{12.89}} & \multicolumn{1}{l}{\textbf{12.25}}\\
             & & \multicolumn{1}{l}
             {Target} &  \multicolumn{1}{l}{13.77} & \multicolumn{1}{l}{13.99} & \multicolumn{1}{l}{\underline{13.12}} & \multicolumn{1}{l}{\textbf{12.38}}\\
         
         & \multirow{2}{*}{ADP} & \multicolumn{1}{l}
         {Predicted} & \multicolumn{1}{l}{{54.02}} & \multicolumn{1}{l}{{37.84}} & \multicolumn{1}{l}{\underline{33.93}} & \multicolumn{1}{l}{\textbf{34.05}}\\
             & & \multicolumn{1}{l}
             {Target} &  \multicolumn{1}{l}{56.68} & \multicolumn{1}{l}{36.09}  & \multicolumn{1}{l}{\underline{31.08}} & \multicolumn{1}{l}{\textbf{32.64}}\\
        
        & \multirow{2}{*}{PIC} & \multicolumn{1}{l}
        {Predicted} &  \multicolumn{1}{l}{13.37} & \multicolumn{1}{l}{23.65}  & \multicolumn{1}{l}{\underline{26.18}} & \multicolumn{1}{l}{\textbf{30.41}}\\
             & & \multicolumn{1}{l}
             {Target} &  \multicolumn{1}{l}{14.97}  & \multicolumn{1}{l}{25.53}   & \multicolumn{1}{l}{\underline{28.97}} & \multicolumn{1}{l}{\textbf{31.75}}\\

        & \multirow{2}{*}{SIC} & \multicolumn{1}{l}
        {Predicted} &  \multicolumn{1}{l}{68.59} & \multicolumn{1}{l}{68.35}  & \multicolumn{1}{l}{\underline{68.92}} & \multicolumn{1}{l}{\textbf{69.68}}\\
             & & \multicolumn{1}{l}
             {Target} &  \multicolumn{1}{l}{68.53}  & \multicolumn{1}{l}{68.26}   & \multicolumn{1}{l}{\underline{70.34}} & \multicolumn{1}{l}{\textbf{70.61}}\\

        & \multirow{2}{*}{AIC} & \multicolumn{1}{l}
        {Predicted} &  \multicolumn{1}{l}{61.34} & \multicolumn{1}{l}{57.92}  & \multicolumn{1}{l}{\underline{62.38}} & \multicolumn{1}{l}{\textbf{64.46}}\\
             & & \multicolumn{1}{l}
             {Target} &  \multicolumn{1}{l}{62.82}  & \multicolumn{1}{l}{60.67}   & \multicolumn{1}{l}{\underline{63.93}} & \multicolumn{1}{l}{\textbf{64.59}}\\
        \midrule
        
    \multirow{16}{*}{ViT-S}
    
        & \multirow{2}{*}{NEG} & \multicolumn{1}{l}
        {Predicted} & \multicolumn{1}{l}{53.29} & \multicolumn{1}{l}{52.81}  & \multicolumn{1}{l}{\underline{55.76}} & \multicolumn{1}{l}{\textbf{56.39}}\\
             & & \multicolumn{1}{l}
             {Target} &  \multicolumn{1}{l}{53.93} & \multicolumn{1}{l}{53.58}  & \multicolumn{1}{l}{\underline{58.71}} & \multicolumn{1}{l}{\textbf{59.46}}\\
        
        & \multirow{2}{*}{POS} & \multicolumn{1}{l}
        {Predicted} &  \multicolumn{1}{l}{14.16} & \multicolumn{1}{l}{14.75}  & \multicolumn{1}{l}{\underline{13.06}} & \multicolumn{1}{l}{\textbf{12.15}}\\
             & & \multicolumn{1}{l}
             {Target} & \multicolumn{1}{l}{13.08} & \multicolumn{1}{l}{14.38}  & \multicolumn{1}{l}{\underline{12.97}} & \multicolumn{1}{l}{\textbf{11.86}}\\
        
        & \multirow{2}{*}{INS} & \multicolumn{1}{l}
        {Predicted} &  \multicolumn{1}{l}{45.72} & \multicolumn{1}{l}{45.21}  & \multicolumn{1}{l}{\underline{46.55}} & \multicolumn{1}{l}{\textbf{47.68}}\\
             & & \multicolumn{1}{l}
             {Target} &  \multicolumn{1}{l}{46.12}  &  \multicolumn{1}{l}{45.69} & \multicolumn{1}{l}{\underline{47.83}} & \multicolumn{1}{l}{\textbf{48.53}}\\
         
         & \multirow{2}{*}{DEL} & \multicolumn{1}{l}
         {Predicted} & \multicolumn{1}{l}{11.28} & \multicolumn{1}{l}{11.92}  & \multicolumn{1}{l}{\underline{11.18}} & \multicolumn{1}{l}{\textbf{10.31}}\\
             & & \multicolumn{1}{l}
             {Target} &  \multicolumn{1}{l}{11.06} & \multicolumn{1}{l}{11.69}  & \multicolumn{1}{l}{\underline{10.98}} & \multicolumn{1}{l}{\textbf{10.16}}\\
             
         & \multirow{2}{*}{ADP} & \multicolumn{1}{l}
         {Predicted} & \multicolumn{1}{l}{51.94} & \multicolumn{1}{l}{36.98} & \multicolumn{1}{l}{\underline{36.74}} & \multicolumn{1}{l}{\textbf{36.40}}\\
             & & \multicolumn{1}{l}
             {Target} &  \multicolumn{1}{l}{50.59} & \multicolumn{1}{l}{64.72} & \multicolumn{1}{l}{\underline{39.58}} & \multicolumn{1}{l}{\textbf{39.43}}\\
        
        & \multirow{2}{*}{PIC} & \multicolumn{1}{l}
        {Predicted} &  \multicolumn{1}{l}{13.67} &  \multicolumn{1}{l}{8.68} & \multicolumn{1}{l}{\underline{15.49}} & \multicolumn{1}{l}{\textbf{17.79}}\\
             & & \multicolumn{1}{l}
             {Target} &  \multicolumn{1}{l}{15.00}  &  \multicolumn{1}{l}{10.02}  & \multicolumn{1}{l}{\underline{18.14}} & \multicolumn{1}{l}{\textbf{19.59}}\\

        & \multirow{2}{*}{SIC} & \multicolumn{1}{l}
        {Predicted} &  \multicolumn{1}{l}{69.46} & \multicolumn{1}{l}{70.19}  & \multicolumn{1}{l}{\underline{70.54}} & \multicolumn{1}{l}{\textbf{72.13}}\\
             & & \multicolumn{1}{l}
             {Target} &  \multicolumn{1}{l}{69.38}  & \multicolumn{1}{l}{72.44}   & \multicolumn{1}{l}{\underline{73.43}} & \multicolumn{1}{l}{\textbf{74.52}}\\

        & \multirow{2}{*}{AIC} & \multicolumn{1}{l}
        {Predicted} &  \multicolumn{1}{l}{63.86} & \multicolumn{1}{l}{64.49}  & \multicolumn{1}{l}{\underline{65.02}} & \multicolumn{1}{l}{\textbf{65.58}}\\
             & & \multicolumn{1}{l}
             {Target} &  \multicolumn{1}{l}{63.45}  & \multicolumn{1}{l}{65.05}   & \multicolumn{1}{l}{\underline{66.89}} & \multicolumn{1}{l}{\textbf{67.62}}\\
        \bottomrule
  \end{tabular}}
  \end{small}
  \end{center}
  \vspace{-2mm}
  \caption{Explanation tests results on the IN dataset (ViT models): For POS, DEL and ADP, lower is better. For NEG, INS, PIC, SIC and AIC, higher is better. See Sec.~\ref{sec:experiments} for details.
  }
  \label{tab:appendix_vit_backbones_metrics}\vspace{-5mm}
\end{table}
\begin{table}
  
   \begin{center}
   \begin{small}
   \scalebox{0.86}{
    \begin{tabular}{@{}lc@{}lc@{}lc@{}lc@{}lc@{}lc@{}lc@{}}
    \toprule
      & & & \multicolumn{1}{l}{GC} & \multicolumn{1}{l}{GC++} & \multicolumn{1}{l}{LIFT} & \multicolumn{1}{l}{AC} & \multicolumn{1}{l}{IIA2} & \multicolumn{1}{l}{IIA3}\\
    \midrule
    \multirow{12}{*}{{IN-SEG}}
    & \multirow{4}{*}{CN}

    & \multicolumn{1}{l}{PA} & \multicolumn{1}{l}{77.01} & \multicolumn{1}{l}{77.54} & \multicolumn{1}{l}{63.77} & \multicolumn{1}{l}{77.04} & \multicolumn{1}{l}{\underline{78.94}} & \multicolumn{1}{l}{\textbf{79.36}}\\

    & & \multicolumn{1}{l}{mAP} & \multicolumn{1}{l}{81.01} & \multicolumn{1}{l}{85.63} & \multicolumn{1}{l}{69.40} & \multicolumn{1}{l}{86.93} & \multicolumn{1}{l}{\underline{87.32}} & \multicolumn{1}{l}{\textbf{88.13}}\\

    & & \multicolumn{1}{l}{mIoU} & \multicolumn{1}{l}{56.58} & \multicolumn{1}{l}{58.35} & \multicolumn{1}{l}{53.81} & \multicolumn{1}{l}{58.42} & \multicolumn{1}{l}{\underline{60.98}} & \multicolumn{1}{l}{\textbf{61.57}}\\
    
    & & \multicolumn{1}{l}{mF1} & \multicolumn{1}{l}{36.88} & \multicolumn{1}{l}{38.26} & \multicolumn{1}{l}{35.91} & \multicolumn{1}{l}{41.29} & \multicolumn{1}{l}{\underline{41.96}} & \multicolumn{1}{l}{\textbf{42.32}}\\
    
    & \multirow{4}{*}{RN}

    & \multicolumn{1}{l}{PA} & \multicolumn{1}{l}{71.93} & \multicolumn{1}{l}{71.96} & \multicolumn{1}{l}{71.68} & \multicolumn{1}{l}{70.36} & \multicolumn{1}{l}{\underline{72.35}} & \multicolumn{1}{l}{\textbf{73.31}}\\

    & & \multicolumn{1}{l}{mAP} & \multicolumn{1}{l}{84.21} & \multicolumn{1}{l}{84.23} & \multicolumn{1}{l}{83.79} & \multicolumn{1}{l}{81.14} & \multicolumn{1}{l}{\underline{84.83}} & \multicolumn{1}{l}{\textbf{85.64}}\\

    & & \multicolumn{1}{l}{mIoU} & \multicolumn{1}{l}{53.06} & \multicolumn{1}{l}{53.29} & \multicolumn{1}{l}{52.17} & \multicolumn{1}{l}{52.91} & \multicolumn{1}{l}{\underline{53.74}} & \multicolumn{1}{l}{\textbf{54.68}}\\
    
    & & \multicolumn{1}{l}{mF1} & \multicolumn{1}{l}{42.51} & \multicolumn{1}{l}{42.68} & \multicolumn{1}{l}{41.95} & \multicolumn{1}{l}{42.08} & \multicolumn{1}{l}{\underline{43.91}} & \multicolumn{1}{l}{\textbf{44.42}}\\

    & \multirow{4}{*}{DN}

    & \multicolumn{1}{l}{PA} & \multicolumn{1}{l}{73.00} & \multicolumn{1}{l}{73.21} & \multicolumn{1}{l}{72.87} & \multicolumn{1}{l}{72.44} & \multicolumn{1}{l}{\textbf{73.64}} & \multicolumn{1}{l}{\underline{73.56}}\\

    & & \multicolumn{1}{l}{mAP} & \multicolumn{1}{l}{85.04} & \multicolumn{1}{l}{85.53} & \multicolumn{1}{l}{84.82} & \multicolumn{1}{l}{84.62} & \multicolumn{1}{l}{\underline{86.03}} & \multicolumn{1}{l}{\textbf{86.19}}\\

    & & \multicolumn{1}{l}{mIoU} & \multicolumn{1}{l}{54.18} & \multicolumn{1}{l}{54.57} & \multicolumn{1}{l}{54.11} & \multicolumn{1}{l}{54.89} & \multicolumn{1}{l}{\underline{55.04}} & \multicolumn{1}{l}{\textbf{55.62}}\\
    
    & & \multicolumn{1}{l}{mF1} & \multicolumn{1}{l}{41.74} & \multicolumn{1}{l}{42.58} & \multicolumn{1}{l}{41.61} & \multicolumn{1}{l}{43.51} & \multicolumn{1}{l}{\underline{43.89}} & \multicolumn{1}{l}{\textbf{44.17}}\\
    
    \midrule
    
    \multirow{12}{*}{COCO}

    & \multirow{4}{*}{CN}
    & \multicolumn{1}{l}{PA} & \multicolumn{1}{l}{68.75} & \multicolumn{1}{l}{66.49} & \multicolumn{1}{l}{60.37} & \multicolumn{1}{l}{64.10} & \multicolumn{1}{l}{\underline{70.38}} & \multicolumn{1}{l}{\textbf{70.73}}\\

    & & \multicolumn{1}{l}{mAP} & \multicolumn{1}{l}{75.02} & \multicolumn{1}{l}{75.21} & \multicolumn{1}{l}{67.98} & \multicolumn{1}{l}{76.09} & \multicolumn{1}{l}{\underline{76.21}} & \multicolumn{1}{l}{\textbf{76.52}}\\

    & & \multicolumn{1}{l}{mIoU} & \multicolumn{1}{l}{43.46} & \multicolumn{1}{l}{44.01} & \multicolumn{1}{l}{37.08} & \multicolumn{1}{l}{44.27} & \multicolumn{1}{l}{\underline{46.48}} & \multicolumn{1}{l}{\textbf{46.90}}\\
    
    & & \multicolumn{1}{l}{mF1} & \multicolumn{1}{l}{28.96} & \multicolumn{1}{l}{29.85} & \multicolumn{1}{l}{26.92} & \multicolumn{1}{l}{30.81} & \multicolumn{1}{l}{\underline{32.45}} & \multicolumn{1}{l}{\textbf{32.56}}\\
    
    & \multirow{4}{*}{RN}

    & \multicolumn{1}{l}{PA} & \multicolumn{1}{l}{64.17} & \multicolumn{1}{l}{64.39} & \multicolumn{1}{l}{64.02} & \multicolumn{1}{l}{63.90} & \multicolumn{1}{l}{\underline{64.89}} & \multicolumn{1}{l}{\textbf{65.77}}\\

    & & \multicolumn{1}{l}{mAP} & \multicolumn{1}{l}{74.19} & \multicolumn{1}{l}{74.27} & \multicolumn{1}{l}{73.78} & \multicolumn{1}{l}{72.80} & \multicolumn{1}{l}{\underline{75.12}} & \multicolumn{1}{l}{\textbf{75.49}}\\

    & & \multicolumn{1}{l}{mIoU} & \multicolumn{1}{l}{42.37} & \multicolumn{1}{l}{43.25} & \multicolumn{1}{l}{42.59} & \multicolumn{1}{l}{42.88} & \multicolumn{1}{l}{\underline{44.56}} & \multicolumn{1}{l}{\textbf{44.93}}\\
    
    & & \multicolumn{1}{l}{mF1} & \multicolumn{1}{l}{31.64} & \multicolumn{1}{l}{32.82} & \multicolumn{1}{l}{31.77} & \multicolumn{1}{l}{32.41} & \multicolumn{1}{l}{\underline{34.95}} & \multicolumn{1}{l}{\textbf{35.03}}\\

    & \multirow{4}{*}{DN}

    & \multicolumn{1}{l}{PA} & \multicolumn{1}{l}{63.50} & \multicolumn{1}{l}{64.06} & \multicolumn{1}{l}{63.25} & \multicolumn{1}{l}{\underline{64.51}} & \multicolumn{1}{l}{\textbf{64.68}} & \multicolumn{1}{l}{64.45}\\

    & & \multicolumn{1}{l}{mAP} & \multicolumn{1}{l}{72.61} & \multicolumn{1}{l}{73.07} & \multicolumn{1}{l}{72.15} & \multicolumn{1}{l}{73.85} & \multicolumn{1}{l}{\underline{74.14}} & \multicolumn{1}{l}{\textbf{74.39}}\\

    & & \multicolumn{1}{l}{mIoU} & \multicolumn{1}{l}{43.02} & \multicolumn{1}{l}{43.75} & \multicolumn{1}{l}{42.85} & \multicolumn{1}{l}{44.16} & \multicolumn{1}{l}{\underline{44.30}} & \multicolumn{1}{l}{\textbf{44.57}}\\
    
    & & \multicolumn{1}{l}{mF1} & \multicolumn{1}{l}{31.04} & \multicolumn{1}{l}{32.31} & \multicolumn{1}{l}{30.83} & \multicolumn{1}{l}{33.93} & \multicolumn{1}{l}{\underline{34.72}} & \multicolumn{1}{l}{\textbf{34.98}}\\
    
     \midrule
    
    \multirow{12}{*}{{VOC}}
    & \multirow{4}{*}{CN}

    & \multicolumn{1}{l}{PA} & \multicolumn{1}{l}{72.54} & \multicolumn{1}{l}{72.09} & \multicolumn{1}{l}{63.32} & \multicolumn{1}{l}{69.83} & \multicolumn{1}{l}{\underline{72.96}} & \multicolumn{1}{l}{\textbf{73.02}}\\

    & & \multicolumn{1}{l}{mAP} & \multicolumn{1}{l}{77.27} & \multicolumn{1}{l}{79.47} & \multicolumn{1}{l}{68.83} & \multicolumn{1}{l}{80.45} & \multicolumn{1}{l}{\underline{80.81}} & \multicolumn{1}{l}{\textbf{82.47}}\\

    & & \multicolumn{1}{l}{mIoU} & \multicolumn{1}{l}{50.28} & \multicolumn{1}{l}{50.63} & \multicolumn{1}{l}{48.86} & \multicolumn{1}{l}{49.76} & \multicolumn{1}{l}{\underline{52.11}} & \multicolumn{1}{l}{\textbf{52.64}}\\
    
    & & \multicolumn{1}{l}{mF1} & \multicolumn{1}{l}{35.24} & \multicolumn{1}{l}{35.67} & \multicolumn{1}{l}{33.26} & \multicolumn{1}{l}{34.51} & \multicolumn{1}{l}{\underline{36.83}} & \multicolumn{1}{l}{\textbf{36.89}}\\
    
    & \multirow{4}{*}{RN}

    & \multicolumn{1}{l}{PA} & \multicolumn{1}{l}{68.74} & \multicolumn{1}{l}{69.01} & \multicolumn{1}{l}{68.61} & \multicolumn{1}{l}{68.00} & \multicolumn{1}{l}{\underline{69.45}} & \multicolumn{1}{l}{\textbf{70.12}}\\

    & & \multicolumn{1}{l}{mAP} & \multicolumn{1}{l}{79.68} & \multicolumn{1}{l}{79.96} & \multicolumn{1}{l}{79.41} & \multicolumn{1}{l}{78.02} & \multicolumn{1}{l}{\underline{80.58}} & \multicolumn{1}{l}{\textbf{81.26}}\\

    & & \multicolumn{1}{l}{mIoU} & \multicolumn{1}{l}{49.44} & \multicolumn{1}{l}{49.91} & \multicolumn{1}{l}{49.15} & \multicolumn{1}{l}{49.32} & \multicolumn{1}{l}{\underline{50.40}} & \multicolumn{1}{l}{\textbf{53.68}}\\
    
    & & \multicolumn{1}{l}{mF1} & \multicolumn{1}{l}{33.08} & \multicolumn{1}{l}{33.56} & \multicolumn{1}{l}{32.69} & \multicolumn{1}{l}{32.74} & \multicolumn{1}{l}{\underline{34.57}} & \multicolumn{1}{l}{\textbf{34.82}}\\

    & \multirow{4}{*}{DN}

    & \multicolumn{1}{l}{PA} & \multicolumn{1}{l}{68.43} & \multicolumn{1}{l}{68.78} & \multicolumn{1}{l}{68.24} & \multicolumn{1}{l}{68.36} & \multicolumn{1}{l}{\underline{69.33}} & \multicolumn{1}{l}{\textbf{70.15}}\\

    & & \multicolumn{1}{l}{mAP} & \multicolumn{1}{l}{78.68} & \multicolumn{1}{l}{79.06} & \multicolumn{1}{l}{78.52} & \multicolumn{1}{l}{78.62} & \multicolumn{1}{l}{\underline{79.96}} & \multicolumn{1}{l}{\textbf{80.27}}\\

    & & \multicolumn{1}{l}{mIoU} & \multicolumn{1}{l}{49.29} & \multicolumn{1}{l}{49.68} & \multicolumn{1}{l}{49.03} & \multicolumn{1}{l}{49.11} & \multicolumn{1}{l}{\underline{50.26}} & \multicolumn{1}{l}{\textbf{50.44}}\\
    
    & & \multicolumn{1}{l}{mF1} & \multicolumn{1}{l}{32.92} & \multicolumn{1}{l}{33.83} & \multicolumn{1}{l}{32.28} & \multicolumn{1}{l}{32.56} & \multicolumn{1}{l}{\underline{34.18}} & \multicolumn{1}{l}{\textbf{34.50}}\\
    \bottomrule
  \end{tabular}}
  \end{small}
  \end{center}
  \vspace{-2mm}
  \caption{Segmentation tests on three datasets (CNN models). For all metrics, higher is better. See Sec.~\ref{sec:experiments} for details.}\vspace{-5mm}
  \label{tab:cnns_segmentation_table}
\end{table}
\begin{table}
  
   \begin{center}
   \begin{small}
   \scalebox{1.05}{
    \begin{tabular}{@{}lc@{}lc@{}lc@{}lc@{}lc@{}}%
    \toprule
      & & & \multicolumn{1}{l}{T-Attr} & \multicolumn{1}{l}{GAE} & \multicolumn{1}{l}{IIA2} & \multicolumn{1}{l}{IIA3}\\
    \midrule
    \multirow{8}{*}{{IN-Seg}}
    & \multirow{4}{*}{ViT-B}
    & \multicolumn{1}{l}{PA} & \multicolumn{1}{l}{79.70} & \multicolumn{1}{l}{76.30} & \multicolumn{1}{l}{\underline{79.80}} & \multicolumn{1}{l}{\textbf{80.71}}\\
    
    & & \multicolumn{1}{l}{mAP} & \multicolumn{1}{l}{86.03} & \multicolumn{1}{l}{85.28} &\multicolumn{1}{l}{\underline{87.27}} & \multicolumn{1}{l}{\textbf{87.38}}\\
    
    & & \multicolumn{1}{l}{mIoU} & \multicolumn{1}{l}{61.95} & \multicolumn{1}{l}{58.34} & \multicolumn{1}{l}{\underline{62.59}} & \multicolumn{1}{l}{\textbf{63.04}}\\
    
    & & \multicolumn{1}{l}{mF1} & \multicolumn{1}{l}{40.17} & \multicolumn{1}{l}{41.85} & \multicolumn{1}{l}{\underline{44.91}} & \multicolumn{1}{l}{\textbf{45.16}}\\
    
    & \multirow{4}{*}{ViT-S}
    & \multicolumn{1}{l}{PA} & \multicolumn{1}{l}{80.86} & \multicolumn{1}{l}{76.66} & \multicolumn{1}{l}{\underline{81.44}}& \multicolumn{1}{l}{\textbf{81.49}}\\
    
    & & \multicolumn{1}{l}{mAP} & \multicolumn{1}{l}{86.13} & \multicolumn{1}{l}{84.23} &
    \multicolumn{1}{l}{\textbf{86.91}}& \multicolumn{1}{l}{\underline{86.85}}\\
    
    & & \multicolumn{1}{l}{mIoU} & \multicolumn{1}{l}{63.61} & \multicolumn{1}{l}{57.70} & \multicolumn{1}{l}{\underline{64.09}} & \multicolumn{1}{l}{\textbf{64.47}}\\
    
    & & \multicolumn{1}{l}{mF1} & \multicolumn{1}{l}{43.60} & \multicolumn{1}{l}{40.72} & \multicolumn{1}{l}{\underline{46.14}} & \multicolumn{1}{l}{\textbf{46.70}}\\
    
    \midrule
    
    \multirow{8}{*}{COCO}
    & \multirow{4}{*}{ViT-B}
    & \multicolumn{1}{l}{PA} & \multicolumn{1}{l}{\underline{68.89}}  & \multicolumn{1}{l}{67.10}  & \multicolumn{1}{l}{68.81} & \multicolumn{1}{l}{\textbf{69.32}}\\
    
    & & \multicolumn{1}{l}{mAP} & \multicolumn{1}{l}{78.57} & \multicolumn{1}{l}{78.72}  &\multicolumn{1}{l}{\underline{80.64}} & \multicolumn{1}{l}{\textbf{81.03}}\\
    
    & & \multicolumn{1}{l}{mIoU} & \multicolumn{1}{l}{46.62} & \multicolumn{1}{l}{46.51} & \multicolumn{1}{l}{\underline{47.75}}  & \multicolumn{1}{l}{\textbf{47.89}}\\
    
    & & \multicolumn{1}{l}{mF1} & \multicolumn{1}{l}{26.28} & \multicolumn{1}{l}{31.70} & \multicolumn{1}{l}{\underline{33.87}} & \multicolumn{1}{l}{\textbf{34.01}}\\

    & \multirow{4}{*}{ViT-S}
    & \multicolumn{1}{l}{PA} & \multicolumn{1}{l}{69.90} & \multicolumn{1}{l}{67.95}  & \multicolumn{1}{l}{\underline{70.31}}& \multicolumn{1}{l}{\textbf{70.60}}\\
    
    & & \multicolumn{1}{l}{mAP} & \multicolumn{1}{l}{79.28} & \multicolumn{1}{l}{78.65} &\multicolumn{1}{l}{\underline{80.53}} & \multicolumn{1}{l}{\textbf{80.89}}\\
    
    & & \multicolumn{1}{l}{mIoU} & \multicolumn{1}{l}{48.62} & \multicolumn{1}{l}{46.52} & \multicolumn{1}{l}{\underline{50.86}}& \multicolumn{1}{l}{\textbf{51.26}}\\
    
    & & \multicolumn{1}{l}{mF1} & \multicolumn{1}{l}{30.88} & \multicolumn{1}{l}{30.96} & \multicolumn{1}{l}{\underline{35.64}} & \multicolumn{1}{l}{\textbf{35.75}}\\
    
     \midrule
    
    \multirow{8}{*}{{VOC}}
    & \multirow{4}{*}{ViT-B}
    & \multicolumn{1}{l}{PA} & \multicolumn{1}{l}{73.70} & \multicolumn{1}{l}{71.32}  & \multicolumn{1}{l}{\underline{75.36}}& \multicolumn{1}{l}{\textbf{75.59}}\\
    
    & & \multicolumn{1}{l}{mAP} & \multicolumn{1}{l}{81.08} & \multicolumn{1}{l}{80.88} &\multicolumn{1}{l}{\textbf{81.96}} & \multicolumn{1}{l}{\underline{81.87}}\\
    
    & & \multicolumn{1}{l}{mIoU} & \multicolumn{1}{l}{53.09} & \multicolumn{1}{l}{51.82} & \multicolumn{1}{l}{\underline{53.64}}  & \multicolumn{1}{l}{\textbf{53.79}}\\
    
    & & \multicolumn{1}{l}{mF1} & \multicolumn{1}{l}{31.50} & \multicolumn{1}{l}{35.72} & \multicolumn{1}{l}{\underline{36.41}} & \multicolumn{1}{l}{\textbf{36.46}}\\
    
    & \multirow{4}{*}{ViT-S}
    & \multicolumn{1}{l}{PA} & \multicolumn{1}{l}{74.96} & \multicolumn{1}{l}{71.85}  & \multicolumn{1}{l}{\underline{76.44}} & \multicolumn{1}{l}{\textbf{76.53}}\\
    
    & & \multicolumn{1}{l}{mAP} & \multicolumn{1}{l}{81.76} & \multicolumn{1}{l}{80.60} &
   \multicolumn{1}{l}{\textbf{82.79}} & \multicolumn{1}{l}{\underline{82.61}}\\
    
    & & \multicolumn{1}{l}{mIoU} & \multicolumn{1}{l}{55.37} & \multicolumn{1}{l}{51.55} & \multicolumn{1}{l}{\textbf{55.92}} & \multicolumn{1}{l}{\underline{55.78}}\\
    
    & & \multicolumn{1}{l}{mF1} & \multicolumn{1}{l}{36.03} & \multicolumn{1}{l}{34.95} & \multicolumn{1}{l}{\textbf{39.33}} & \multicolumn{1}{l}{\underline{39.26}}\\
    \bottomrule
  \end{tabular}}
  \end{small}
  \end{center}\vspace{-2mm}
  \caption{Segmentation tests on three datasets (ViT models).}
  \label{tab:vit_segmentation}
\end{table}

\subsection{Results}
\label{subsec:results}
\paragraph{Explanation Tests}
\label{subsubsection:obj2}

\begin{figure}[t]
\centering
    \includegraphics[width=0.5\textwidth]{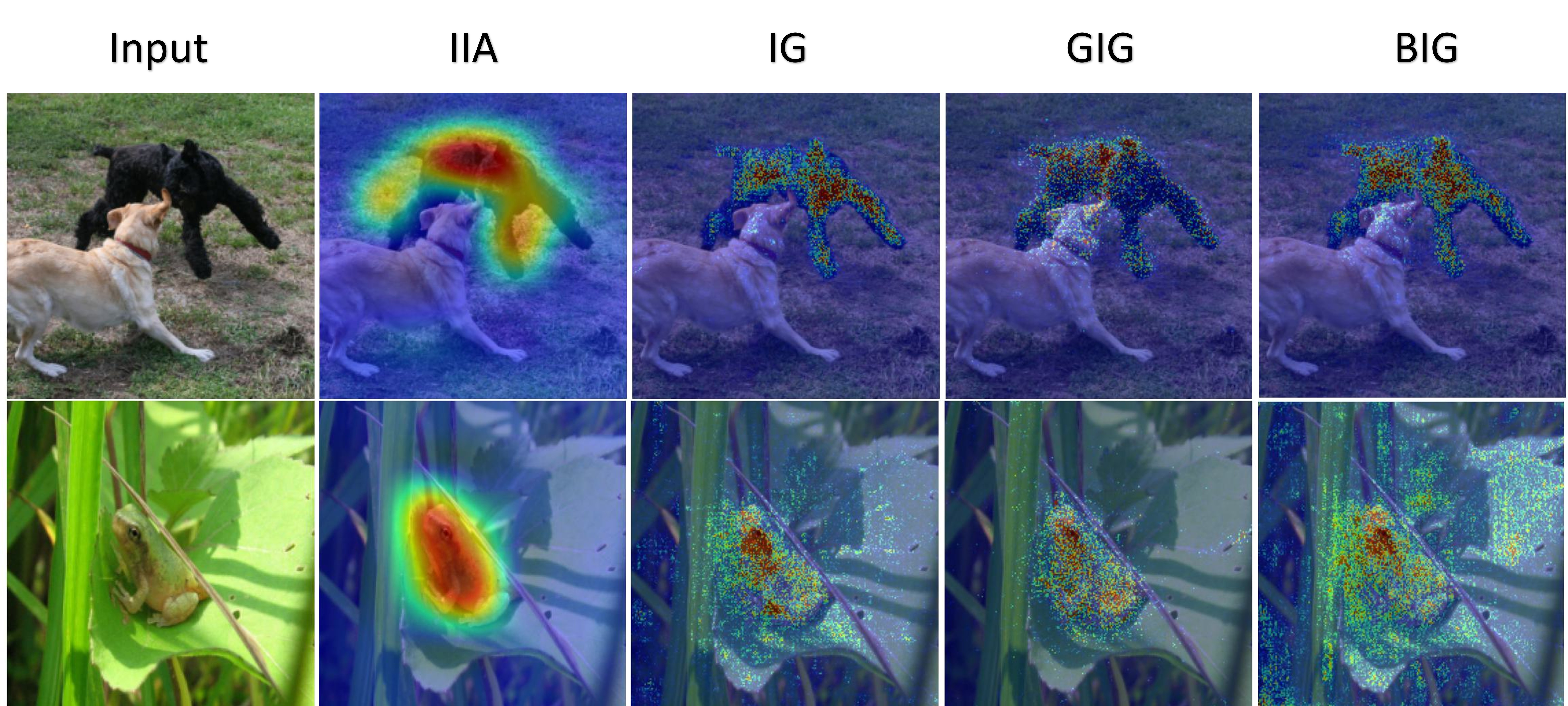}
    \caption{Explanation maps produced for IIA (IIA3) and three path integral baselines using CN w.r.t. the `Kerry blue terrier' (top) and 'tailed frog, bell toad, ribbed toad, tailed toad, Ascaphus trui' (bottom) classes.
    }\vspace{-3mm}
    \label{fig:igtrends}
\end{figure}

Tables~\ref{tab:cnn_backbones_metrics_exp} and \ref{tab:appendix_vit_backbones_metrics} present explanation tests for CNN and ViT models, respectively. The results encompass all combinations of datasets, models, explanation methods, explanation metrics, and settings (target and predicted). Notably, IIA consistently outperforms all baselines across all metrics and architectures. Among the IIA variants, IIA3 surpasses IIA2 for both ViTs (Tab.~\ref{tab:appendix_vit_backbones_metrics}) and CNNs (Tab.~\ref{tab:cnn_backbones_metrics_exp}, holding true for the vast majority of model-metric combinations). This trend underscores the advantage of triple integration, which incorporates information from layer $L-1$. In CNNs, GC and GC++ are the runner up, utilizing both activation and gradients, and outperforming other methods across most metrics. Furthermore, path integration methods (IG, BIG, and GIG) exhibit competitive results on POS and DEL metrics but demonstrate weaker performance on other metrics. This divergence could be attributed to the granular output maps generated by integration-based methods, as depicted in Fig.~\ref{fig:igtrends}. These methods focus solely on integration within the input space, ignoring activations, and thus might overlook key features. Notably, achieving high performance on POS and poor performance on NEG metrics reinforces this observation. As path integration methods yield sparse maps that may impact performance in certain metrics, we also report results for the SIC and AIC metrics~\cite{kapishnikov2019xrai}, employed in the evaluation of GIG\cite{kapishnikov2021guided} and BIG\cite{xu2020attribution}. However, the inclusion of SIC and AIC metrics does not alter the observed trends in the results. This finding emphasizes that IIA is exceptionally effective in generating high-quality explanation maps. 
\vspace{-3mm}

\paragraph{Segmentation Tests}
Tables~\ref{tab:cnns_segmentation_table} and \ref{tab:vit_segmentation} present segmentation tests results on CNN and ViT models, respectively. The results are reported for all combinations of datasets, models, explanation methods, and segmentation metrics. For these experiments, we exclusively consider the top 5 performing CNN explanation methods from Tab.~\ref{tab:cnn_backbones_metrics_exp}. Once again, it is evident that IIA is the best performer, yielding the most accurate segmentation results for both CNN and ViT models. %

\begin{figure}
\centering
    \includegraphics[width=0.5\textwidth]{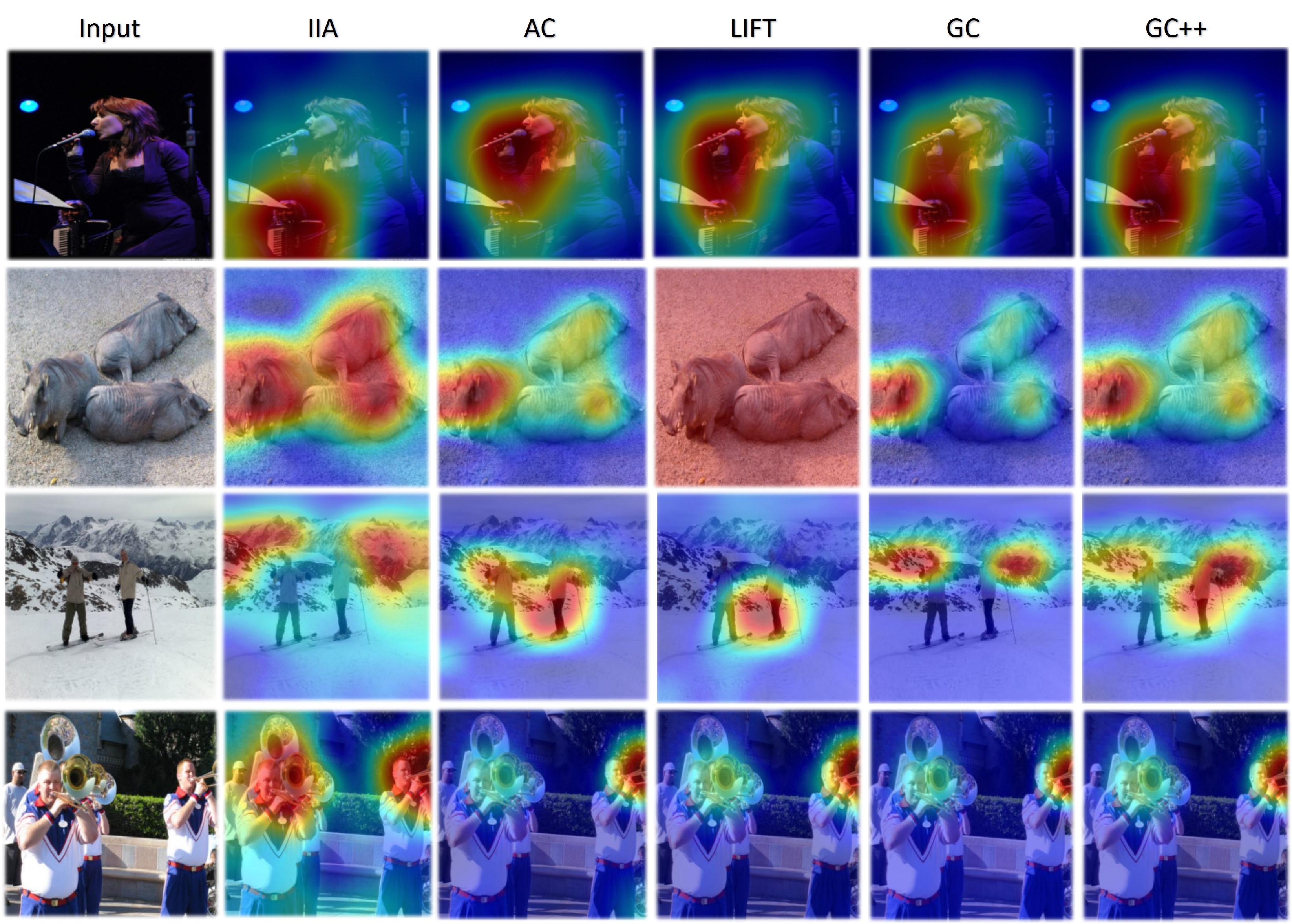}
    \caption{Qualitative Results: Explanation maps produced using ConvNext w.r.t. the classes (top to bottom): `accordion, piano accordion, squeeze box', `warthog', `alp', and `trombone'.
    }\vspace{-4mm}
    \label{fig:qualitative_fig}
\end{figure}

\begin{figure}
\centering
    \includegraphics[width=0.48\textwidth]{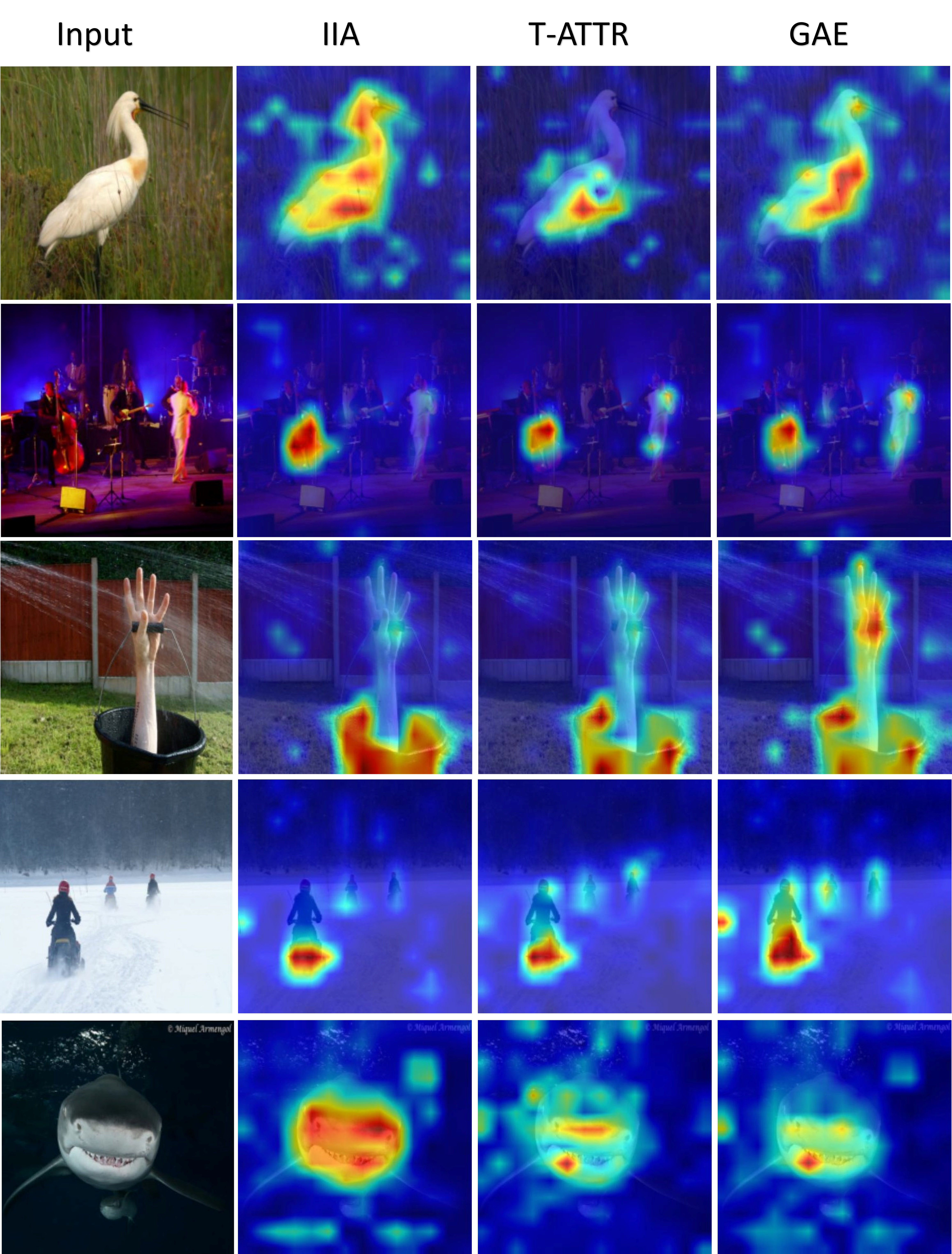}
    \caption{Qualitative Results: Explanation maps produced using ViT-B w.r.t. the classes (top to bottom): `spoonbill', `cello, violoncello', 'bucket, pail', `snowmobile', and `tiger shark'.}
   \vspace{-3mm}\label{fig:qrvit}
\end{figure}

\vspace{-3mm}
\paragraph{Qualitative Evaluation}
Figures~\ref{fig:qualitative_fig} and \ref{fig:qrvit} present a qualitative comparison of the explanation maps obtained by the top-performing CNN explanation methods and ViT explanation methods, respectively. These examples are randomly selected from multiple classes within the IN dataset. Arguably, IIA (IIA3) produces the most accurate explanation maps in terms of class discrimination and localization. These results align well with the trends observed in Tabs.~\ref{tab:cnn_backbones_metrics_exp}-\ref{tab:vit_segmentation}. For example, in Fig.~\ref{fig:qualitative_fig}, IA distinguishes itself by capturing multiple objects related to the target class, setting it apart from the other methods. We further observe that in the case of class `accordion, piano accordion, and squeeze box', IIA focuses mostly on the correct item, while the gradient-free methods like AC and LIFT focus mostly on different parts of the image, showcasing their class-agnostic behavior. Interestingly, in the second row, LIFT generates a flat explanation map, a phenomenon warranting further investigation in future research. Additional qualitative results for both CNN and ViT models are provided in Appendix~\ref{sec:more-qual}.

\subsection{Ablation Study}
\label{sec:ablation-study}
In this work, we employ IIA with double and triple integrals. In this section, we investigate the contribution and necessity of these choices. To this end, we consider three alternatives: (1) \textbf{IMG} - only the input image is interpolated, i.e., we set $b_0=1$ and $b_j=0$ for all $j>0$. (2) \textbf{ACT} - only the representation (activation or attention maps) produced by the $L$-th layer in the model is interpolated, i.e., we set $b_L=1$ and $b_j=0$ for all $j<L$. Note that for both IMG and ACT, we set $l=L$ in Eq.~\ref{eq:IIA-vanilla-approx}, i.e., the integrand is computed w.r.t. the $L$-th layer. (3) \textbf{IIA2 (L-1)} - performs double integral, but interpolates on the layer $L-1$ instead of the last layer $L$ (by setting $b_0=1$, $b_{L-1}=1$, and $b_j=0$ for all other layers).

Table~\ref{tab:appendix_vit_ablation} reports the results for the RN and ViT-B models on the IN dataset under the target settings. For the sake of completeness, we further include the results for IG, IIA2, and IIA3 (Tabs.~\ref{tab:cnn_backbones_metrics_exp} and \ref{tab:appendix_vit_backbones_metrics}). We see that IIA2 and IIA3 perform the best. While ACT is inferior to IIA2, it outperforms IMG. This underscores the need to interpolate on the activations. Yet, the contributions from both IMG and ACT are complementary, as can be seen in IIA2 that combines both.

\begin{figure}
\centering
    \includegraphics[width=0.5\textwidth]{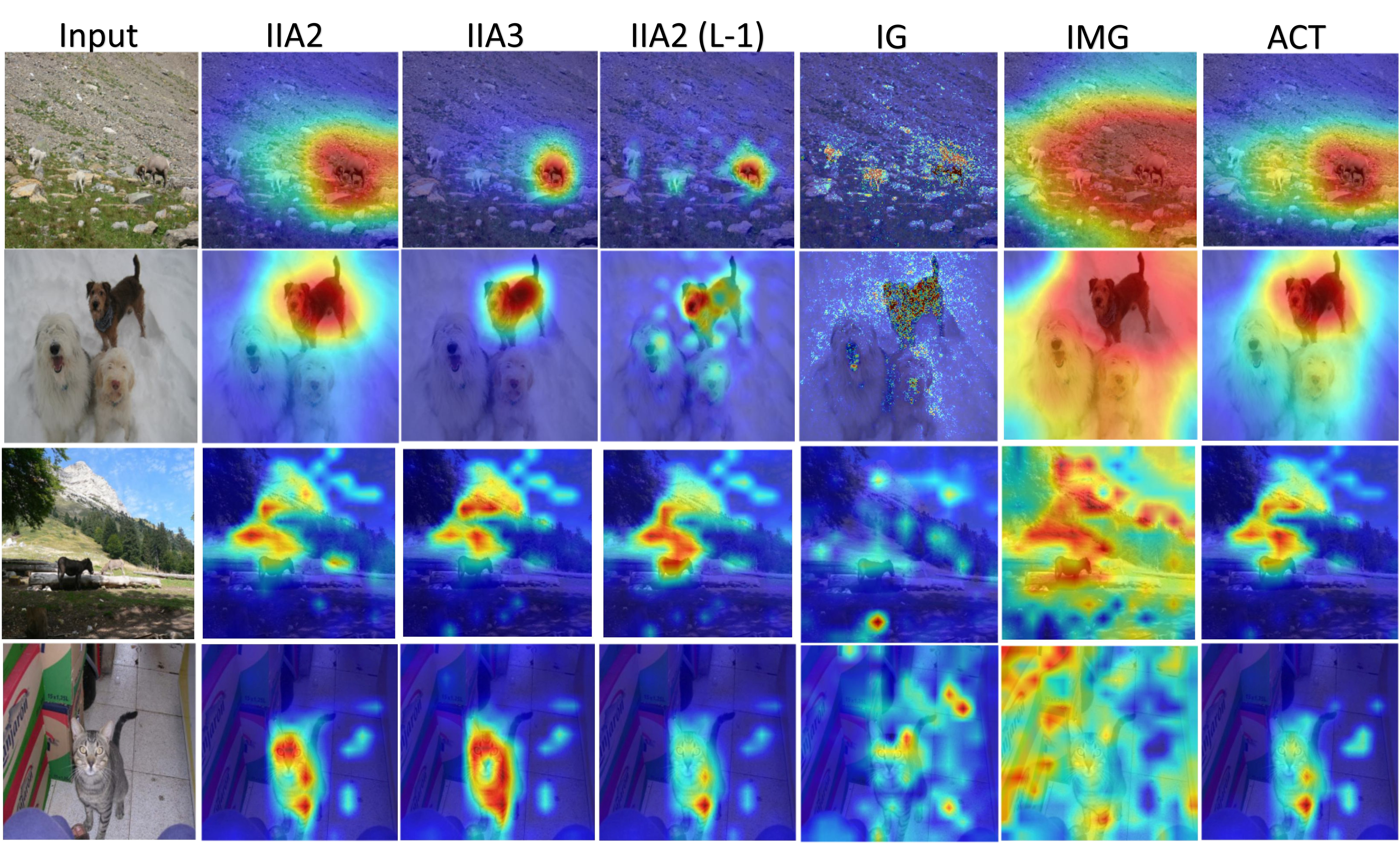}
    \caption{Explanation maps produced using RN (rows 1,2) and ViT (rows 3,4) w.r.t. the classes (top to bottom): `bighorn, bighorn sheep, cimarron, Rocky Mountain bighorn, Rocky Mountain sheep, Ovis canadensis','Irish terrier', 'alp', 'Egyptian cat'.
    }\vspace{-3mm}
    \label{fig:iiaablation}
\end{figure}

Interestingly, IIA2 (L-1) outperforms IIA2 and IIA3 in terms of POS and DEL metrics, on the RN model. Figure~\ref{fig:iiaablation} demonstrates this trend visually. This finding suggests that IIA2 (L-1) generates more focused maps as it utilizes the penultimate layer, which has a higher spatial feature map resolution of $14\times14$ (compared to $7\times7$ in the last convolutional layer in RN), hence is capable of producing more focused explanation maps that lead to better performance on POS and DEL metrics. This is due to the fact that the deletion of the most relevant pixels results in fewer pixels being removed, and the mask is more focused on a subset of pixels compared to IIA2. IIA2 (that operates on the last layer with lower resolution)  produces less focused explanation maps that may highlight irrelevant areas. Such coarse highlighting leads to a slower decrease in the prediction score during the deletion process. Yet, on all other metrics (except POS and DEL) IIA2 (L-1) is inferior to IIA2. Moreover, in the case of ViT, where the resolution is fixed across all layers (as all layers output the same number of token representations), IIA2 outperforms IIA2 (L-1) across the board. Thus, we conclude that under the same spatial resolution, the last layer (both in RN and ViT) enables better feature aggregation than the penultimate layer.

\begin{table}[t!]
  
  \begin{center}
  \begin{small}
  \scalebox{0.93}{
    \begin{tabular}{@{}lc@{}lc@{}lc@{}lc@{}}%
    \toprule
      & &  \multicolumn{1}{l}{IMG} & \multicolumn{1}{l}{ACT} & \multicolumn{1}{l}{IG} & \multicolumn{1}{l}{IIA2 (L-1)} & \multicolumn{1}{l}{IIA2} & \multicolumn{1}{l}{IIA3} \\
    \midrule
    \multirow{8}{*}{RN}

       & \multirow{1}{*}{NEG} & \multicolumn{1}{l}{51.33} & \multicolumn{1}{l}{53.74} & \multicolumn{1}{l}{42.02} & \multicolumn{1}{l}{52.07} & \multicolumn{1}{l}{\underline{56.22}} & \multicolumn{1}{l}{\textbf{56.89}}\\

        & \multirow{1}{*}{POS} &   \multicolumn{1}{l}{19.94} & \multicolumn{1}{l}{21.76}& \multicolumn{1}{l}{16.93} & \multicolumn{1}{l}{\textbf{12.61}} & \multicolumn{1}{l}{16.69} & \multicolumn{1}{l}{\underline{15.78}}\\

        & \multirow{1}{*}{INS} &  \multicolumn{1}{l}{44.54} & \multicolumn{1}{l}{45.38}& \multicolumn{1}{l}{37.55} & \multicolumn{1}{l}{46.82} & \multicolumn{1}{l}{\underline{48.05}} & \multicolumn{1}{l}{\textbf{48.53}}\\

        & \multirow{1}{*}{DEL}  & \multicolumn{1}{l}{{15.28}} & \multicolumn{1}{l}{{14.36}} & \multicolumn{1}{l}{13.46} & \multicolumn{1}{l}{\textbf{10.32}} & \multicolumn{1}{l}{12.82}& \multicolumn{1}{l}{\underline{12.16}}\\

         & \multirow{1}{*}{ADP} & \multicolumn{1}{l}{{17.21}} & \multicolumn{1}{l}{{15.30}} & \multicolumn{1}{l}{36.51} & \multicolumn{1}{l}{38.46} & \multicolumn{1}{l}{\textbf{12.31}} & \multicolumn{1}{l}{\underline{12.40}}\\

        & \multirow{1}{*}{PIC}  &  \multicolumn{1}{l}{35.12} & \multicolumn{1}{l}{39.59}  & \multicolumn{1}{l}{21.43}& \multicolumn{1}{l}{21.08} &  \multicolumn{1}{l}{\textbf{45.21}}& \multicolumn{1}{l}{\underline{45.06}} \\

        & \multirow{1}{*}{SIC}  &  \multicolumn{1}{l}{72.79} & \multicolumn{1}{l}{75.37}  & \multicolumn{1}{l}{51.54}& \multicolumn{1}{l}{72.26}  & \multicolumn{1}{l}{\underline{78.13}} & \multicolumn{1}{l}{\textbf{79.94}}\\

        & \multirow{1}{*}{AIC}  &  \multicolumn{1}{l}{68.87} & \multicolumn{1}{l}{71.29}  & \multicolumn{1}{l}{52.71}& \multicolumn{1}{l}{67.28} & \multicolumn{1}{l}{\underline{75.88}} & \multicolumn{1}{l}{\textbf{76.59}}\\
           
        \midrule

    \multirow{8}{*}{ViT-B}
    
        & \multirow{1}{*}{NEG} & \multicolumn{1}{l}{48.15} & \multicolumn{1}{l}{46.43} & \multicolumn{1}{l}{40.94} & \multicolumn{1}{l}{56.52} & \multicolumn{1}{l}{\underline{57.47}} & \multicolumn{1}{l}{\textbf{58.31}}\\

        & \multirow{1}{*}{POS} &   \multicolumn{1}{l}{19.40} & \multicolumn{1}{l}{22.83} & \multicolumn{1}{l}{22.43} & \multicolumn{1}{l}{17.78}  & \multicolumn{1}{l}{\underline{15.81}} & \multicolumn{1}{l}{\textbf{15.02}}\\

        & \multirow{1}{*}{INS} &  \multicolumn{1}{l}{45.86} & \multicolumn{1}{l}{41.66} & \multicolumn{1}{l}{35.07} & \multicolumn{1}{l}{50.24} & \multicolumn{1}{l}{\underline{50.49}} & \multicolumn{1}{l}{\textbf{51.26}}\\

        & \multirow{1}{*}{DEL}  & \multicolumn{1}{l}{{16.31}} & \multicolumn{1}{l}{{18.19}} & \multicolumn{1}{l}{17.90} & \multicolumn{1}{l}{14.76}  & \multicolumn{1}{l}{\underline{13.12}}& \multicolumn{1}{l}{\textbf{12.38}}\\

         & \multirow{1}{*}{ADP} & \multicolumn{1}{l}{{34.39}} & \multicolumn{1}{l}{{39.62}} & \multicolumn{1}{l}{41.35}  & \multicolumn{1}{l}{38.19} & \multicolumn{1}{l}{\textbf{31.08}} & \multicolumn{1}{l}{\underline{32.64}}\\

        & \multirow{1}{*}{PIC}  &  \multicolumn{1}{l}{25.78} & \multicolumn{1}{l}{22.90}  & \multicolumn{1}{l}{16.89} & \multicolumn{1}{l}{25.64}  & \multicolumn{1}{l}{\underline{28.97}} & \multicolumn{1}{l}{\textbf{31.75}}\\

        & \multirow{1}{*}{SIC}  &  \multicolumn{1}{l}{68.83} & \multicolumn{1}{l}{69.16}   & \multicolumn{1}{l}{58.91} & \multicolumn{1}{l}{69.22} & \multicolumn{1}{l}{\underline{70.34}} & \multicolumn{1}{l}{\textbf{70.61}}\\

        & \multirow{1}{*}{AIC}  &  \multicolumn{1}{l}{62.86} & \multicolumn{1}{l}{63.51} & \multicolumn{1}{l}{54.93} & \multicolumn{1}{l}{63.42} & \multicolumn{1}{l}{\underline{63.93}} & \multicolumn{1}{l}{\textbf{64.59}}\\
        \bottomrule
  \end{tabular}}
  \end{small}
  \end{center}\vspace{-2mm}
  \caption{Ablation study results on the IN dataset (Sec.~\ref{sec:ablation-study}).
  }\vspace{-2mm}
  \label{tab:appendix_vit_ablation}
\end{table}

\section{Conclusion}
\label{sec:conclusion}
We introduced Iterated Integrated Attributions (IIA) - a universal machinery for generating explanations for vision models. IIA employs iterative accumulation of information from interpolated internal network representations and their gradients. Our experiments highlight IIA's effectiveness in explaining both CNN and ViT models, consistently outperforming state-of-the-art explanation methods across diverse tasks, datasets, models, and metrics.

{\small
\bibliographystyle{iccv23.bst}
\bibliography{iccv23}
}

\newpage
\appendix
\onecolumn
\textbf{\Large{Supplementary Materials for Visual Explanations via Iterated Integrated Attributions}}\\

\section{Evaluation Metrics}
\label{sec:metric_desc}

There is no single measure or test set which is generally acceptable for evaluating explanation maps. Hence, in order to ensure comparability, the evaluations in this research  follow earlier works~\cite{chattopadhay2018grad,8354201,chefer2021transformer,kapishnikov2019xrai,petsiuk2018rise}. In general, the various tests entail different types of masking of the original input according to the explanation maps and investigating the change in the model's prediction for the masked input compared to its original prediction based on the unmasked input. There are two variants for these tests which differ based on the class of reference. In one variant, the difference in predictions refers to the ground-truth class, and in the second variant, the difference in predictions refers to the model's original top-predicted class. In the manuscript, we report results for both variants and dub the first variant as `target' and the second variant as `predicted', respectively. 

In what follows, we list and define the different evaluation measures used in this research:
\begin{enumerate}
   \item {Average Drop Percentage \textbf{(ADP)}~\cite{chattopadhay2018grad}: $\text{ADP}=100\%\cdot\frac{1}{N} \sum_{i=1}^N \frac{\max(0,Y_i^c-O_i^c)}{Y_i^c},$ where $N$ is the total number of images in the evaluated dataset, $Y_i^c$ is the model's output score (confidence) for class $c$ w.r.t. the original image $i$. $O_i^c$ is the same model’s score, this time w.r.t. to a masked version of the original image (produced by the Hadamard product of the original image with the explanation map). The \textbf{lower} the ADP the better the result.}
   \item {Percentage of Increase in Confidence \textbf{(PIC)}~\cite{chattopadhay2018grad}: $\text{PIC}=100\%\cdot\frac{1}{N} \sum_{i=1}^N \mathbbm{1}(Y_i^c<O_i^c)$. PIC reports the percentage of cases in which the model’s output scores increase as a result of the replacement of the original image with the masked version based on the explanation map. The explanation map is expected to mask the background and help the model to focus on the original image. Hence, the \textbf{higher} the PIC the better the result.}
   \item Perturbation tests entail a stepwise process in which pixels in the original image are gradually masked out according to their relevance score obtained from the explanation map~\cite{chefer2021transformer}. At each step, an additional 10\% of the pixels are removed and the original image is gradually blacked out. The performance of the explanation model is assessed by measuring the area under the curve (AUC) with respect to the model's prediction on the masked image compared to its prediction with respect to the original (unmasked) image. 

   We consider two types of masking:
   \begin{enumerate}
       \item {Positive perturbation (\textbf{POS}), in which we mask the pixels in decreasing order, from the highest relevance to the lowest, and expect to see a steep decrease in performance, indicating that the masked pixels are important to the classification score. Hence, for the POS perturbation test, lower values indicate better performance.}
       \item {Negative perturbation (\textbf{NEG}), in which we mask the pixels in increasing order, from lowest to highest. A good explanation would maintain the accuracy of the model while removing pixels that are not related to the class of interest. Hence, for the NEG perturbation test, lower values indicate better performance.}
   \end{enumerate}
   {In both positive and negative perturbations, we measure the area-under-the-curve (AUC), for erasing between \text{10\%-90\%} of the pixels.
As explained above, results are reported with respect to the `predicted' or the `target' (ground-truth) class.}
   \item{The deletion and insertion metrics ~\cite{petsiuk2018rise} are described as follows:}
   \begin{enumerate}
       \item {The deletion (\textbf{DEL}) metric measures a decrease in the probability of the class of interest as more and more important pixels are removed, where the importance of each pixel is obtained from the generated explanation map. A sharp drop and thus a low area under the probability curve (as a function of the fraction of removed pixels) means a good explanation.}
       \item {In contrast, the insertion (\textbf{INS}) metric measures the increase in probability as more and more pixels are revealed, with higher AUC indicative of a better explanation.}
   \end{enumerate}
   {Note that there are several ways in which pixels can be removed from an image ~\cite{dabkowski2017real}. In this work, we remove pixels by setting their value to zero. Gradual removal or introduction of pixels is performed in steps of 0.1 i.e., remove or introduce 10\% of the pixels on each step).}

   \item {The Accuracy Information Curve (\textbf{AIC}) and the Softmax Information Curve (\textbf{SIC})~\cite{kapishnikov2019xrai} metrics are both similar in spirit to the receiver operating characteristics (ROC). These measures are inspired by the Bokeh effect in photography~\cite{liu2016stereo}, which consists of focusing on objects of interest while keeping the rest of the image blurred. In a similar fashion, we start with a completely blurred image and gradually sharpen the image areas that are deemed important by a given explanation method. Gradually sharpening the image areas increases the information content of the image. We then compare the explanation methods by measuring the approximate image entropy (e.g., compressed image size) and the model’s performance (e.g., model accuracy).

}
    \begin{enumerate}
       \item { 
       
The AIC metric measures the accuracy of a model as a function of the amount of information provided to the explanation method. AIC is defined as the AUC of the accuracy vs. information plot. The information provided to the method is quantified by the fraction of input features that are considered during the explanation process.

}
       \item { 
       
The SIC metric measures the information content of the output of a softmax classifier as a function of the amount of information provided to the explanation method. SIC is defined as the AUC of the entropy vs. information plot. The entropy of the softmax output is a measure of the uncertainty or randomness of the classifier's predictions. The information provided to the method is quantified by the fraction of input features that are considered during the explanation process.

}
   \end{enumerate}
 \end{enumerate}

\section{Baselines Description}
\label{sec:base_desc}

\begin{enumerate}
       \item {Grad-CAM (\textbf{GC})~\cite{selvaraju2017grad} 

       integrates the activation maps from the last convolutional layer in the CNN by employing global average pooling on the gradients and utilizing them as weights for the feature map channels. 
       }
       \item {Grad-CAM++ (\textbf{GC++})~\cite{chattopadhay2018grad} 

       is an advanced variant of Grad-CAM that utilizes a weighted average of the pixel-wise gradients to generate the activation map weights.
       }
       \item {XGrad-CAM (\textbf{XGC})~\cite{Fu2020AxiombasedGT} 

       calculates activation coefficients using two axioms. Although the authors derived coefficients that satisfy these axioms as closely as possible, their derivation is only demonstrated for ReLU-CNNs.
       }
       \item {Integrated Gradients (\textbf{IG})~\cite{SundararajanTY17} integrates over the interpolated image gradients. }
       \item {Blur IG (\textbf{BIG})~\cite{xu2020attribution} is concerned with the introduction of information using a baseline and opts to use a path that progressively removes Gaussian blur from the attributed image. }
       \item {Guided IG (\textbf{GIG})~\cite{kapishnikov2021guided} improves upon Integrated Gradients by introducing the idea of an adaptive path method. By calculating integration along a different path than Integrated Gradients, high gradient areas are avoided which often leads to an overall reduction in irrelevant attributions.}
       \item {LIFT-CAM (\textbf{LIFT})~\cite{Jung2021TowardsBE} employs the DeepLIFT~\cite{Shrikumar2017LearningIF} technique to estimate the activation maps SHAP values~\cite{Lundberg2017AUA} and then combine them with the activation maps to produce the explanation map.}
       \item {The FullGrad (\textbf{FG}) method~\cite{Srinivas2019FullGradientRF} provides a complete modeling approach of the gradient by also taking the gradient with respect to the bias term, and not just with respect to the input.}
       \item {LayerCAM (\textbf{LC})~\cite{jiang2021layercam} 
      
       utilizes both gradients and activations, but instead of using the Grad-CAM approach and applying pooling on the gradients, it treats the gradients as weights for the activations by assigning each location in the activations with an appropriate gradient location. The explanation map is computed with a location-wise product of the positive gradients (after ReLU) with the activations, and the map is then summed w.r.t. the activation channel, with a ReLU applied to the result.
       }
       \item {Ablation-CAM (\textbf{AC})~\cite{Desai2020AblationCAMVE} 
       
       is an approach that only uses the channels of the activations. It takes each activation channel, masks it from the final map by zeroing out all locations of this channel in the explanation map produced by all the channels, computes the score on the masked explanation map (the map without the specific channel), and this score is used to assign an importance weight for every channel. At last, a weighted sum of the channels produces the final explanation map.
       }
       \item {The Transformer attribution (\textbf{T-ATTR})~\cite{chefer2021transformer} method computes the importance of each input token by analyzing the attention weights assigned to it during self-attention. Specifically, it computes the relevance score of each token as the sum of its attention weights across all layers of the Transformer. The intuition behind this approach is that tokens that receive more attention across different layers are likely more important for the final prediction.
To obtain a more interpretable and localized visualization of the importance scores, the authors also propose a variant of the method called Layer-wise Relevance Propagation (LRP), which recursively distributes the relevance scores back to the input tokens based on their contribution to the intermediate representations.}
       \item {Generic Attention Explainability (\textbf{GAE}) ~\cite{chefer2021generic} is a generalization of T-Attr for explaining Bi-Modal transformers.}
\end{enumerate}

\section{Sanity Checks for Explanation Maps}
\label{sec:res_sanity}
In order to further evaluate the soundness and validity of IIA, we conducted both the \emph{parameter randomization} and \emph{data randomization} sanity tests as proposed by~\cite{adebayo2018sanity}.

Unless stated otherwise, the experiments utilize the ImageNet ILSVRC 2012 validation set~\cite{imagenet} with the VGG-19~\cite{vgg} model and IIA3.

\subsection{Parameter Randomization Test}
\label{subsec:modelrand_test}
The parameter randomization test compares the explanation maps produced by the explanation method based on two setups of the same model architecture: (1) trained - the model is trained on the dataset (e.g., a pretrained VGG-19 model that was trained on ImageNet, and (2) random - the same model architecture, with random weights (e.g., a randomly initialized VGG-19 model). For a method that relies on the actual model to be explained, we anticipate significant differences in the explanation maps produced for the trained model and those produced for the random model. Conversely, if the explanation maps are similar, we conclude that the explanation method is insensitive to the model's parameters, and thus may not be useful for explaining and debugging the model. 

Given a trained model, we consider two types of parameter randomization tests: The first test randomly re-initializes all weights of the model in a cascading fashion (layer after layer). The second test independently randomizes one layer at a time, while keeping all other layers fixed. In both cases, we compare the resulting explanations obtained by using the model with random weights to those derived from the original weights of the model.

\begin{figure}[t]
    \centering
    \includegraphics[scale=0.95]{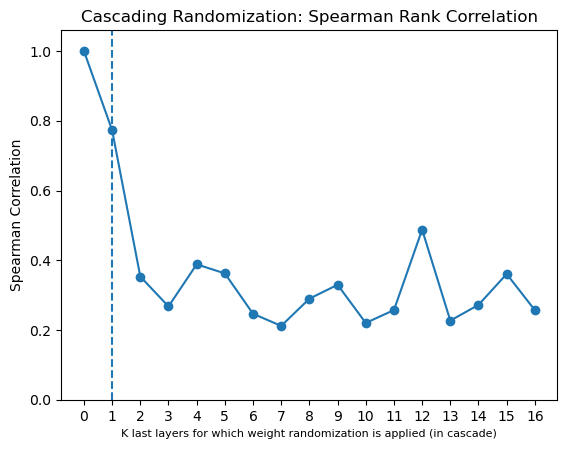}
    \caption{\textbf{Cascading Randomization}: 
  The VGG-19 model is subjected to successive weights randomization, beginning from the last model's layers on the ImageNet dataset. The presented graph depicts the Spearman rank correlation (averaged on 50K examples) between the explanation produced by IIA using the original and randomized model's weights. The x-axis corresponds to the number of layers being randomized, starting from the output layer. The dashed line indicates the point where the successive randomization of the network commences, which is at the top layer. The first dot (x=0) corresponds to no randomization (the original model is used), hence the correlation between the explanation maps is perfect. See Sec.~\ref{subsubsection:cascading_randomization} for further details.}
    \label{fig:sanity_checkes_metric_cascade}
\end{figure}

\paragraph{Cascading Randomization}
\label{subsubsection:cascading_randomization}
The cascading randomization method involves the randomization of a model's weights, starting from the top layer and successively moving down to the bottom layer. This process leads to the destruction of the learned weights from the top to the bottom layers.
 Figure~\ref{fig:sanity_checkes_metric_cascade} presents the Spearman correlation (averaged on 50K examples) between the original explanation map obtained by IIA and the original (pretrained) VGG-19 model and the explanation map obtained by IIA and each of the cascade randomization versions of the original model. The markers on the x-axis are between '0' and '16', where $x=k$ means that the weights of the last $k$ layers of the model are randomized. At $x=0$ there is no randomization, hence the correlation with the original model is perfect. Starting from $x=1$ (marked by the horizontal dashed line) and up to $x=16$, the graph depicts a progressive cascade randomization of the original model. We observe that as more layers' weights are randomized, the correlation with the explanation map of the original model significantly deteriorates. This behavior showcases the sensitivity of IIA to the model's parameters -  an expected and desired property for any explanation method~\cite{adebayo2018sanity}.

\begin{figure}[t]
    \centering
    \includegraphics[scale=0.30]{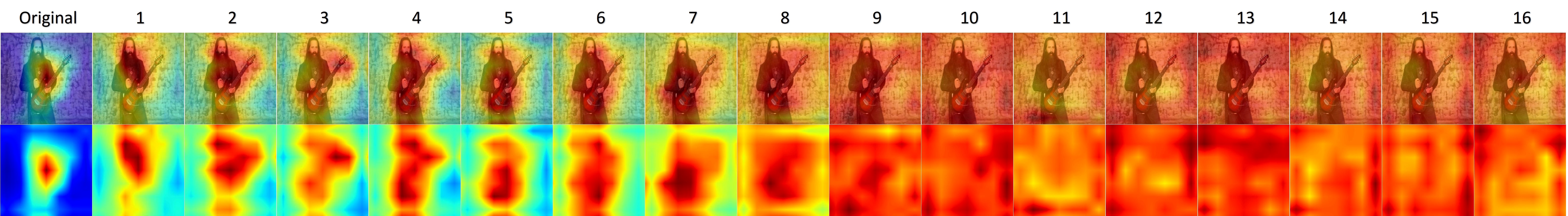}
    \caption{\textbf{Cascading Randomization on VGG-19 (ImageNet)}: The figure presents the original explanations (first column) for `electric guitar'. Progression from left to right depicts the gradual randomization of network weights up to the layer number depicted at the top of the column (starting from the last layer). See Sec.~\ref{subsubsection:cascading_randomization} for further details.}
    \label{fig:sanity_checkes_visual_cascade}
\end{figure}

Figure~\ref{fig:sanity_checkes_visual_cascade} displays a representative example of explanation maps (bottom) and their overlay to the original image (top), illustrating the cascading randomization process. The first column presents explanation maps produced by IIA and the original model, while the rest of the columns present explanation maps produced by IIA and cascading randomized models, where the number $i$ above each column indicates that the explanation map is produced by a model in which the weights of the last $i$ layers were randomized. It is evident that the quality of produced explanation maps significantly degrades as more and more layers are set with random weights.

\paragraph{Independent Randomization}
\label{subsubsection:independ_randomization}

We further consider another version of the model's parameters randomization test, in which a layer-by-layer randomization is employed, one layer at a time. In this test, we aim to isolate the influence of the randomization of each layer, hence randomization is applied to one layer's weights at a time, while all other layers' weights are kept identical to their values in the original model. This randomization methodology enables comprehensive evaluation of the sensitivity of the explanation maps w.r.t. each of the model's layers.

\begin{figure}[t]
    \centering
    \includegraphics[scale=0.95]{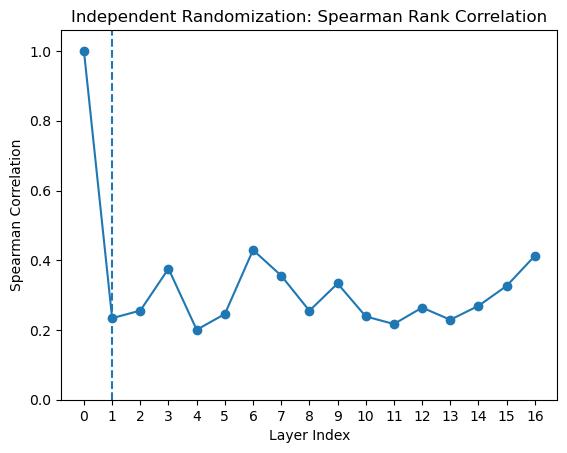}
    \caption{\textbf{Independent Randomization}: 
The randomization process is carried out independently for each layer of the model, while the remaining weights are retained at their pretrained values. The y-axis of the presented graph represents the rank correlation between the original and randomized explanations, with each point on the x-axis corresponding to a specific layer of the model. The dashed line marks the point where the randomization of the network layers commences, which is at the top layer.}
\label{fig:sanity_checkes_metric_ind}
\end{figure}

\begin{figure}[t]
    \centering
    \includegraphics[scale=0.30]{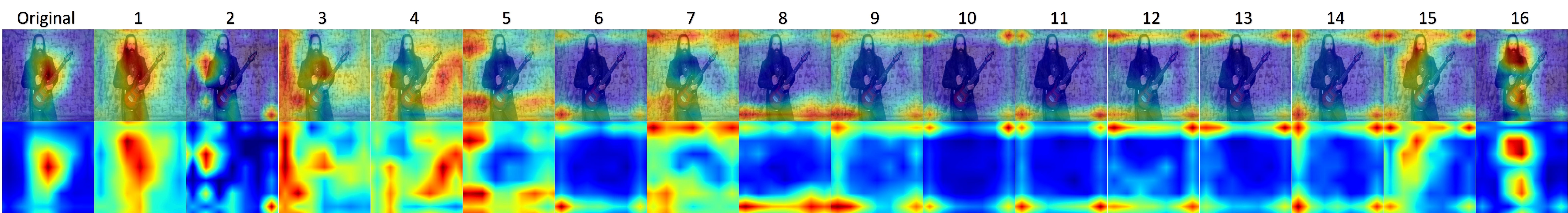}
    \caption{\textbf{Independent Randomization on VGG-19 (ImageNet)}: Similar to Fig.~\ref{fig:sanity_checkes_visual_cascade}, however, this time, each specific layer is randomized independently, while the rest of the weights are kept at their pretrained values.}
    \label{fig:sanity_checkes_visual_ind}
\end{figure}

Figure~\ref{fig:sanity_checkes_metric_ind} presents results for the independent randomization tests. 
At $x=0$ no randomization was applied and the correlation to the original model is perfect. For $x=i$ ($i>0$) the graph indicates the correlation of the original model with a model in which only the weights of the  $i$-th penultimate layer were randomized while the weights of all other layers were kept untouched.  We observe that the correlation values are low across all layers which indicates IIA's sensitivity to weight randomization in each layer separately. This property is a desired property for an explanation method, as it indicates the method's sensitivity to each of the model's layers, independently. 
Finally, Fig.~\ref{fig:sanity_checkes_visual_ind} presents a qualitative example in the same fashion as Fig.~\ref{fig:sanity_checkes_visual_cascade}, this time for the independent randomization test. We observe that the quality of all explanation maps produced by a randomized version of the model differs significantly from the original explanation map. We conclude that IIA successfully passes both types of parameter randomization tests.

\subsection{Data Randomization Test}
\label{subsec:datarand_test}

\begin{figure}[t]
    \centering
    \includegraphics[scale=0.8]{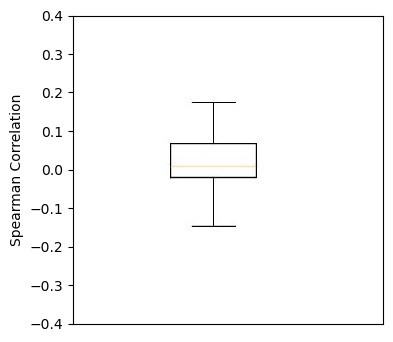}
    \caption{\textbf{Data Randomization Test}: Spearman rank correlation box plot for IIA with the VGG-19 model.}
    \label{fig:sanity_checkes_visual_bxplt}
\end{figure}

The data randomization sanity test is a method used to assess whether an explanation method is sensitive to the labeling of the data used for training the model. This is done by comparing the explanation maps produced by the explanation method for two models with identical architecture that were trained on two different datasets: one with the original labels and another with randomly permuted labels. If the explanation method is sensitive to the labeling of the dataset, we would expect the produced explanation maps to differ significantly between the two cases. However, if the method is insensitive to the permuted labels, it indicates that it does not depend on the relationship between instances and labels that exists in the original data. To conduct the data randomization test, we permute the training labels in the dataset and train the model to achieve a training set accuracy greater than 95\%. Note that the resulting model's test accuracy is never better than randomly guessing a label. We then compute explanations on the same test inputs for both the model trained on true labels and the model trained on randomly permuted labels. Figure~\ref{fig:sanity_checkes_visual_bxplt} presents a box plot computed for the Spearman correlation values obtained for paired explanation maps (50K examples): one produced using the original model that is trained with the ground truth, and another produced by the model trained with the permuted labels. 
We can see that the correlation values are very low indicating IIA's sensitivity to the labeling of the dataset. Hence, we conclude that IIA successfully passes the data randomization test.

\begin{figure}[t]
    \centering
    \includegraphics[scale=0.95]{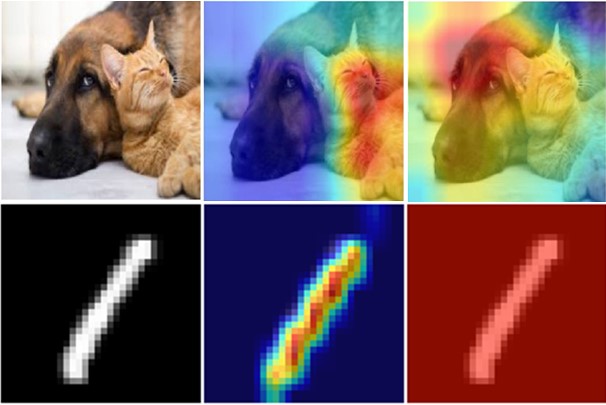}
    \caption{Sanity checks. Rows 1 and 2 present IIA results for the \emph{parameter randomization} and \emph{data randomization} tests w.r.t. the ``tabby cat'' (ImageNet) and ``one'' (MNIST) classes, using ResNet50 and LeNet-5, respectively. Left to right: Row 1: Original image, explanation map produced by IIA and the trained model, explanation map produced by IIA and untrained model (model's weights are randomly initialized without further training). Row 2: Original image, explanation map produced by IIA and a model trained with the ground truth labels, explanation map produced by IIA and a model trained with random labels.}
    \label{fig:sanity_checkes}
\end{figure}

Finally, Figure~\ref{fig:sanity_checkes} presents additional qualitative examples for both tests, this time with different models. The first row shows two explanation maps produced by IIA w.r.t. the ``tabby cat'' class. We see that when IIA utilizes an ImageNet pretrained ResNet50 model, it produces a focused explanation map (around the cat), but when applying IIA to the same model with random weights, it fails to detect the cat in the image. The second row shows that IIA produces an adequate explanation map when the model (LeNet-5~\cite{lecun1998gradient}) is trained with the MNIST ground truth labels but fails when the model is trained with random labels.

\section{Gradient Rollout Implementation}
\label{sec:gr}
The Gradient Rollout (\textbf{GR}) technique is a modified version of the Attention Rollout (\textbf{AR})~\cite{abnar2020quantifying} method, which differentiates itself by including a Hadamard product between each attention map and its gradients in the computation, rather than relying solely on the attention map.
The GR method can be expressed mathematically as follows:
\begin{equation}
\label{eq:gr1}
A'_b=I+E_{h}(A_b \circ G_b),
\end{equation}
\begin{equation}
\label{eq:gr2}
GR=A'_1 \cdot A'_2 \cdot \cdot \cdot A'_B.
\end{equation}
where $A_b$ is a 3D tensor consisting of the 2D attention maps produced by each attention head in the transformer block $b$, $G_b$ is the gradients w.r.t. $A_b$.
$I$ is the identity matrix, $B$ is the number of transformer blocks in the model, $E_h$ is the mean reduction operation (taken across the attention heads dimension), and $\circ$ and $\cdot$ are the Hadamard product and matrix multiplication operators, respectively.

\section{Additional Qualitative Results}
\label{sec:more-qual}
\begin{figure*}
\centering
    \includegraphics[width=0.80\textwidth, height=0.90\textheight]{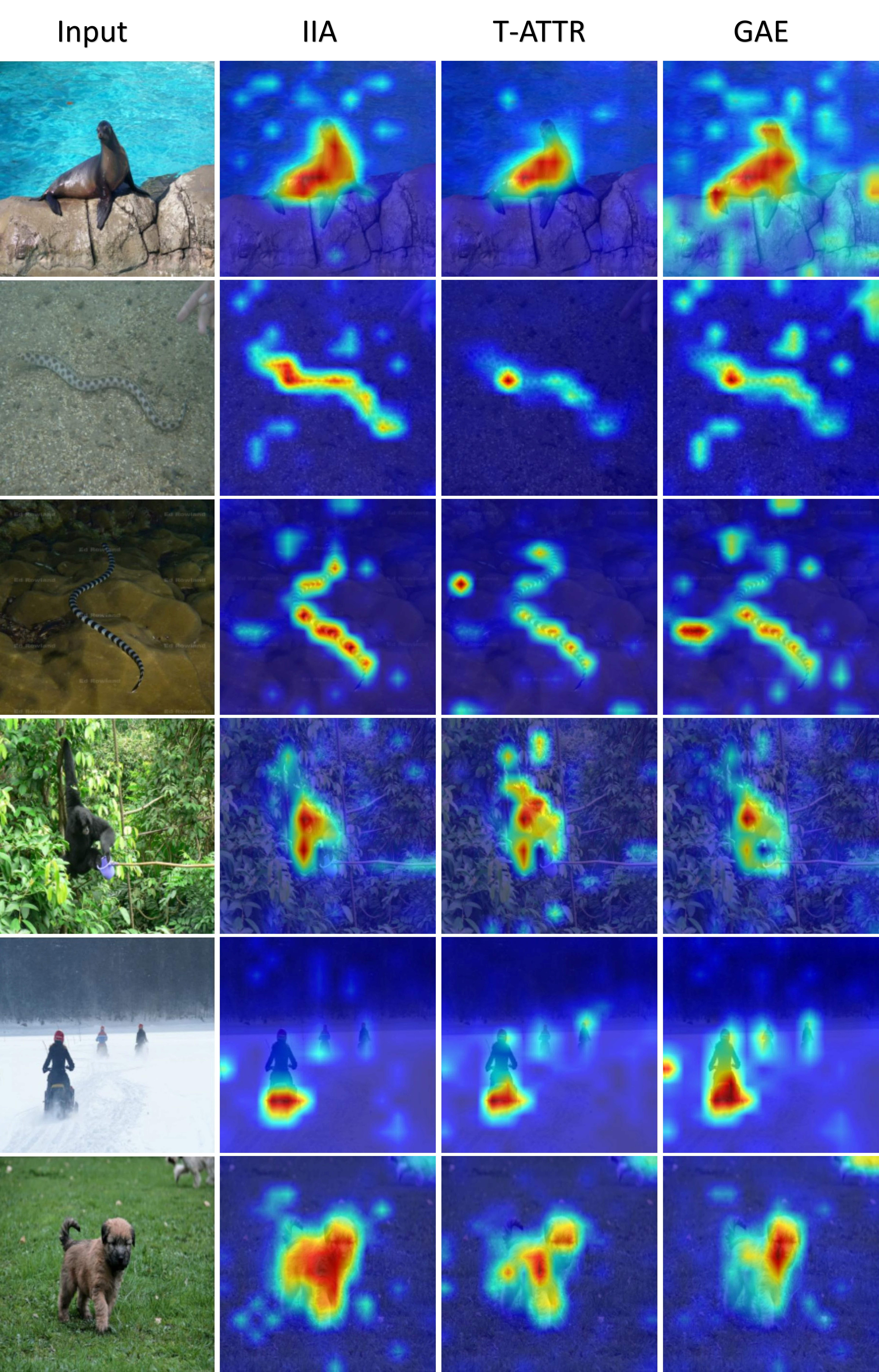}
    
    \caption{Visualizations obtained by explanation methods for ViT-B model. The ground-truth labels of the images are listed according to the format '($\langle$row\#$\rangle$) $\langle$class names$\rangle$': (1) 'sea lion', (2-3) 'sea snake', (4) 'siamang, Hylobates syndactylus, Symphalangus syndactylus', (5) 'snowmobile', (6) 'soft-coated wheaten terrier'.}
    \label{fig:qrv1}
\end{figure*}

\begin{figure*}
\centering
    \includegraphics[width=0.80\textwidth, height=0.90\textheight]{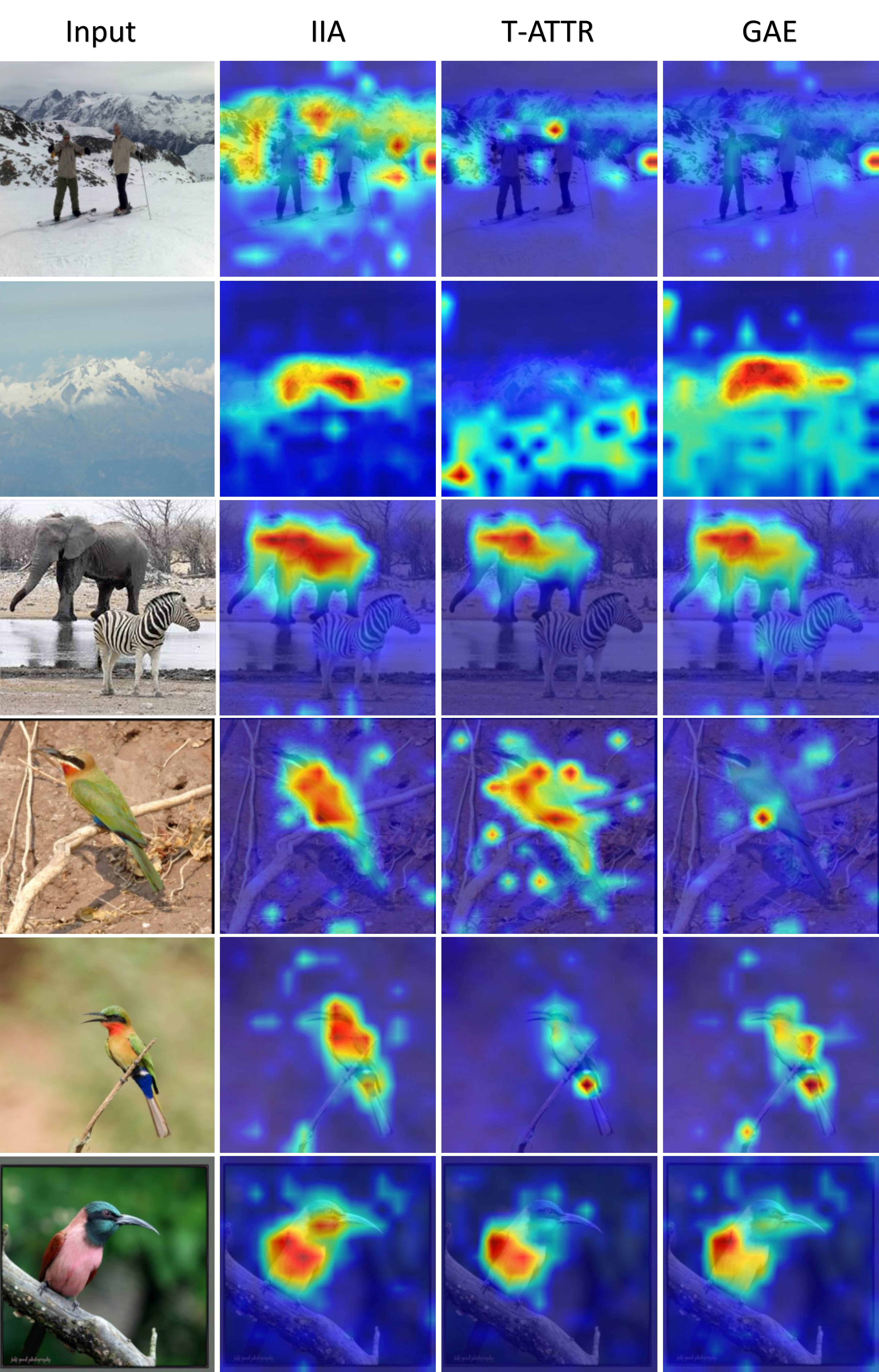}
    
    \caption{Visualizations obtained by explanation methods for ViT-B model. The ground-truth labels of the images are listed according to the format '($\langle$row\#$\rangle$) $\langle$class names$\rangle$': (1-2) 'alp', (3)’Indian elephant, Elephas maximus’, (4-6) 'bee eater'.}
    \label{fig:qrv2}
\end{figure*}

\begin{figure*}
\centering
    \includegraphics[width=0.80\textwidth, height=0.90\textheight]{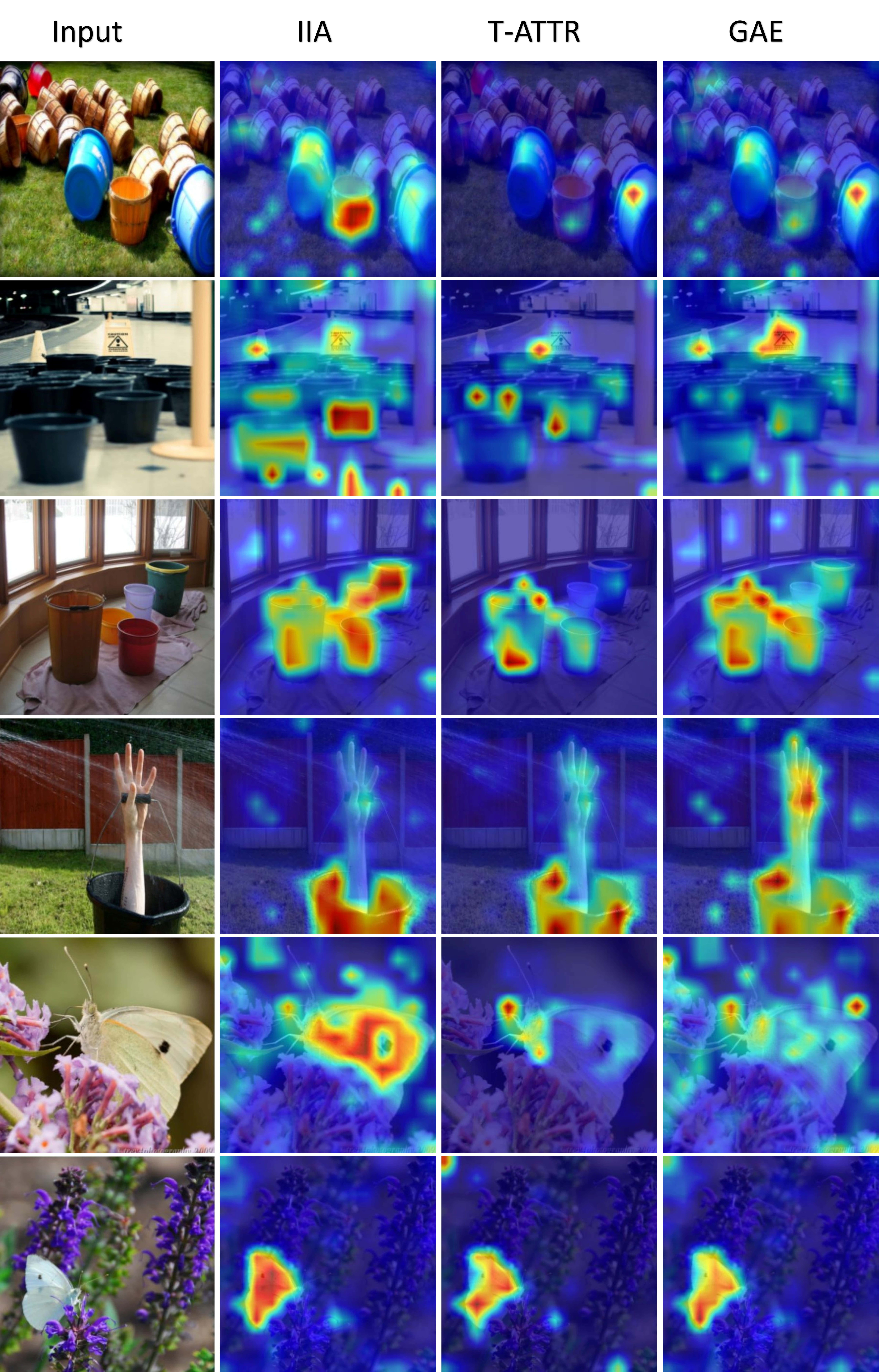}
    
    \caption{Visualizations obtained by explanation methods for ViT-B model. The ground-truth labels of the images are listed according to the format '($\langle$row\#$\rangle$) $\langle$class names$\rangle$': (1-4) 'bucket, pail', (5-6) 'cabbage butterfly'.}
    \label{fig:qrv3}
\end{figure*}

\begin{figure*}
\centering
    \includegraphics[width=0.80\textwidth, height=0.90\textheight]{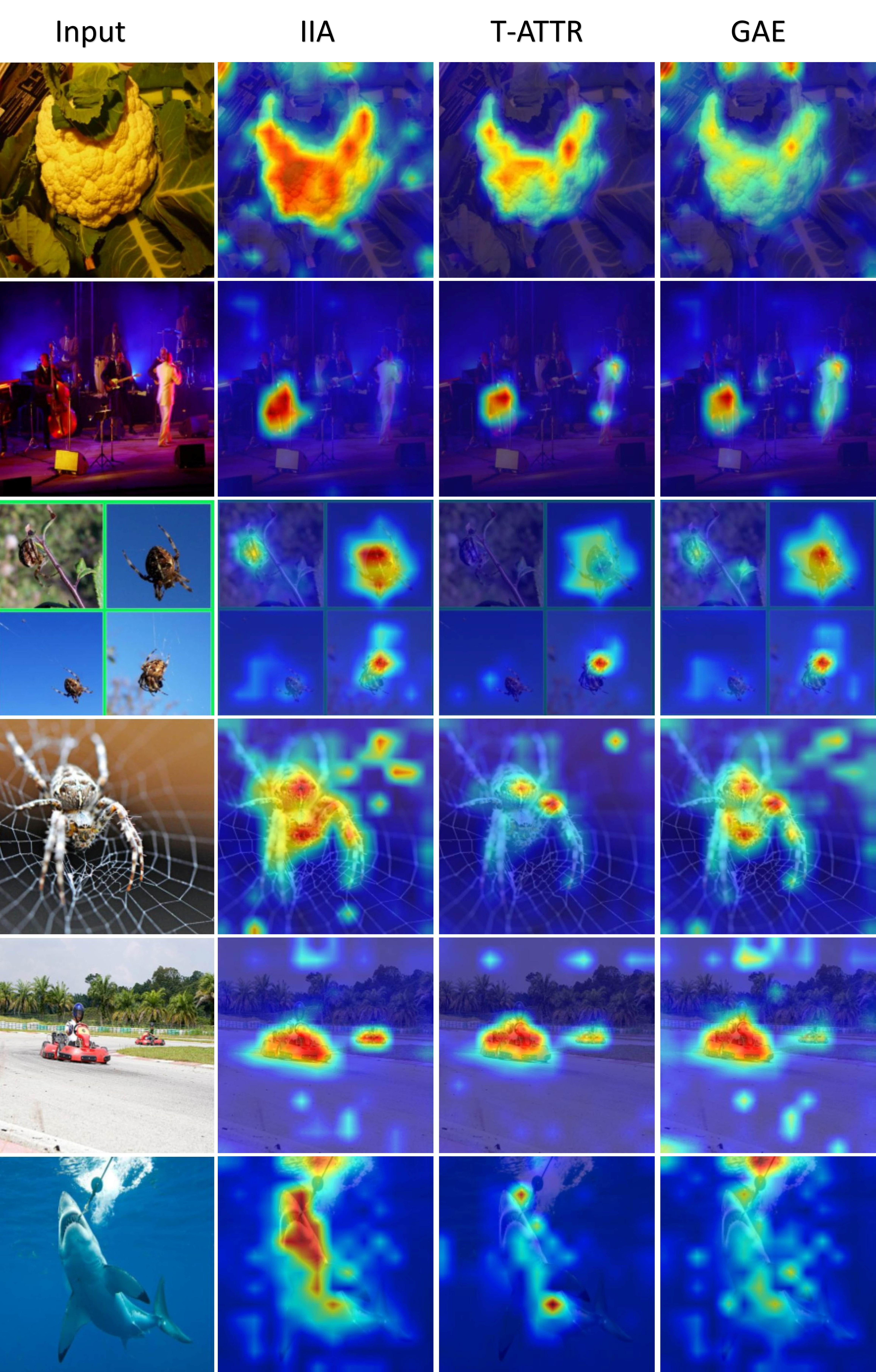}
    
    \caption{Visualizations obtained by explanation methods for ViT-B model. The ground-truth labels of the images are listed according to the format '($\langle$row\#$\rangle$) $\langle$class names$\rangle$': (1) 'cauliflower', (2) 'cello, violoncello', (3-4) 'garden spider, Aranea diademata', (5) 'go-kart', (6) 'great white shark, white shark, man-eater, man-eating shark, Carcharodon carcharias'.}
    \label{fig:qrv4}
\end{figure*}

\begin{figure*}
\centering
    \includegraphics[width=0.80\textwidth, height=0.90\textheight]{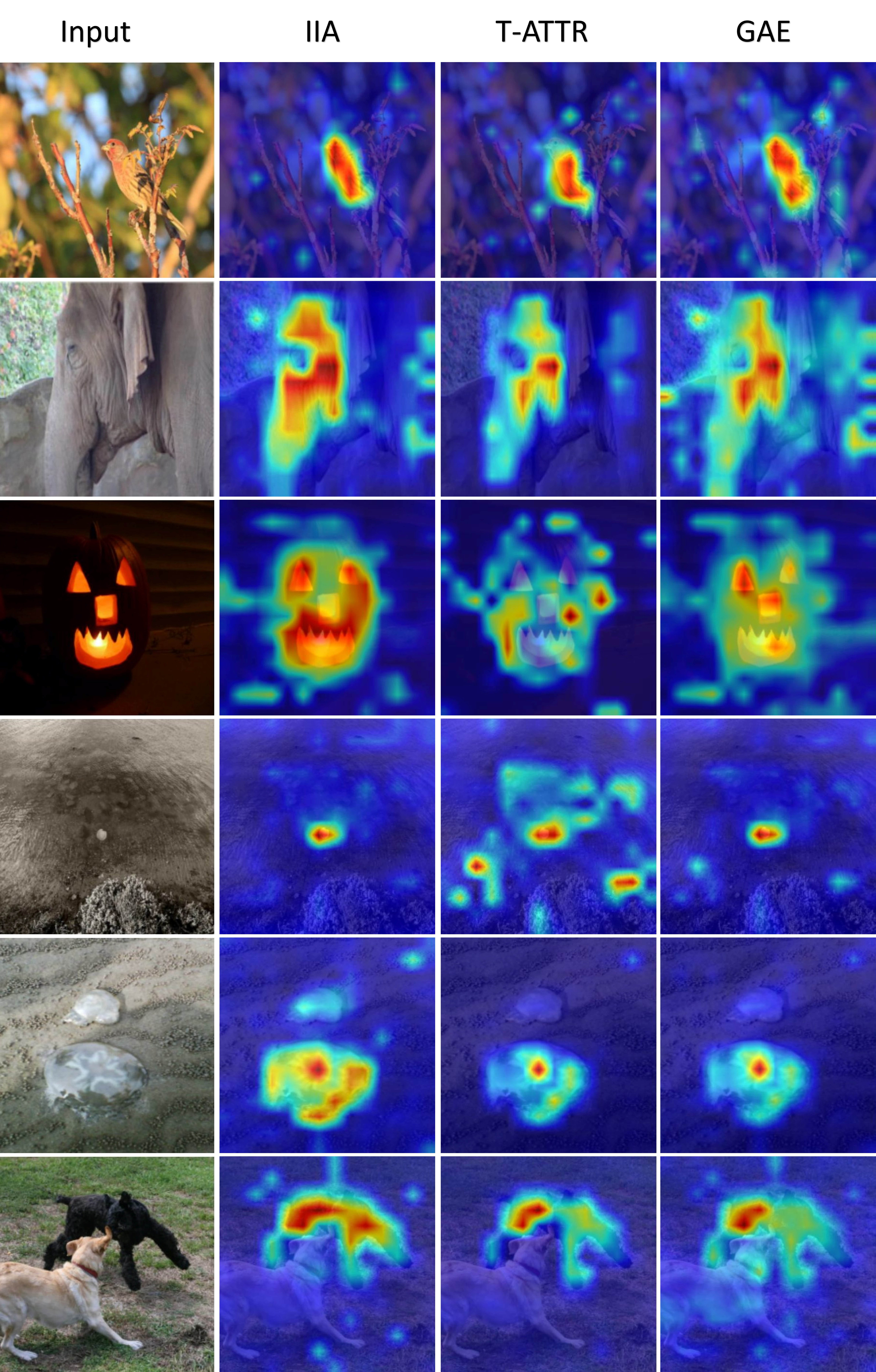}
    
    \caption{Visualizations obtained by explanation methods for ViT-B model. The ground-truth labels of the images are listed according to the format '($\langle$row\#$\rangle$) $\langle$class names$\rangle$': (1) 'house finch, linnet, Carpodacus mexicanus', (2) 'Indian elephant, Elephas maximus', (3) 'jack-o'-lantern', (4-5) 'jellyfish', (6) 'Kerry blue terrier'.}
    \label{fig:qrv5}
\end{figure*}

\begin{figure*}
\centering
    \includegraphics[width=0.80\textwidth, height=0.90\textheight]{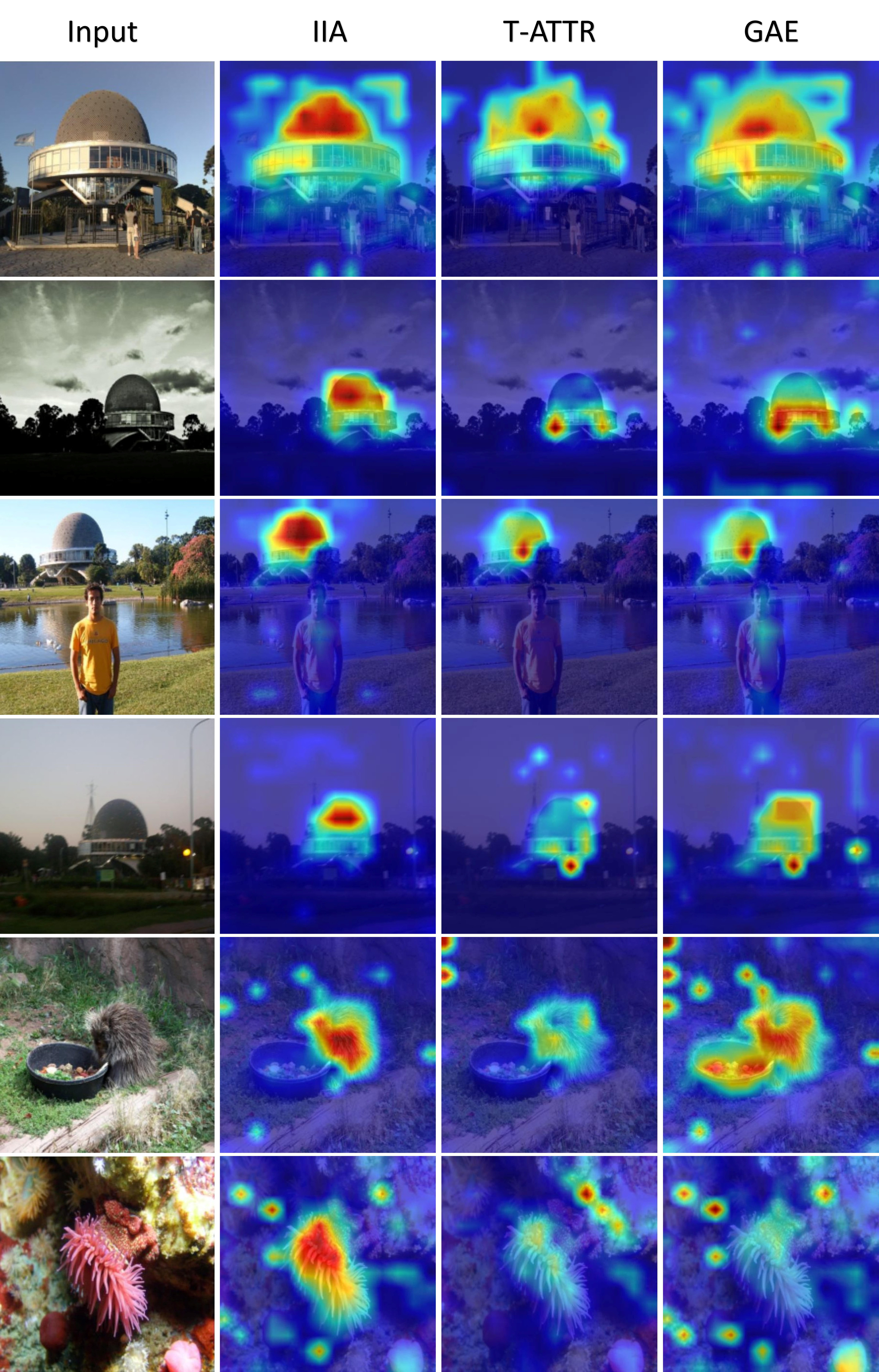}
    
    \caption{Visualizations obtained by explanation methods for ViT-B model. The ground-truth labels of the images are listed according to the format '($\langle$row\#$\rangle$) $\langle$class names$\rangle$': (1-4) 'planetarium', (5) 'porcupine, hedgehog', (6) 'sea anemone, anemone'.}
    \label{fig:qrv6}
\end{figure*}

\begin{figure*}
\centering
    \includegraphics[width=0.80\textwidth, height=0.90\textheight]{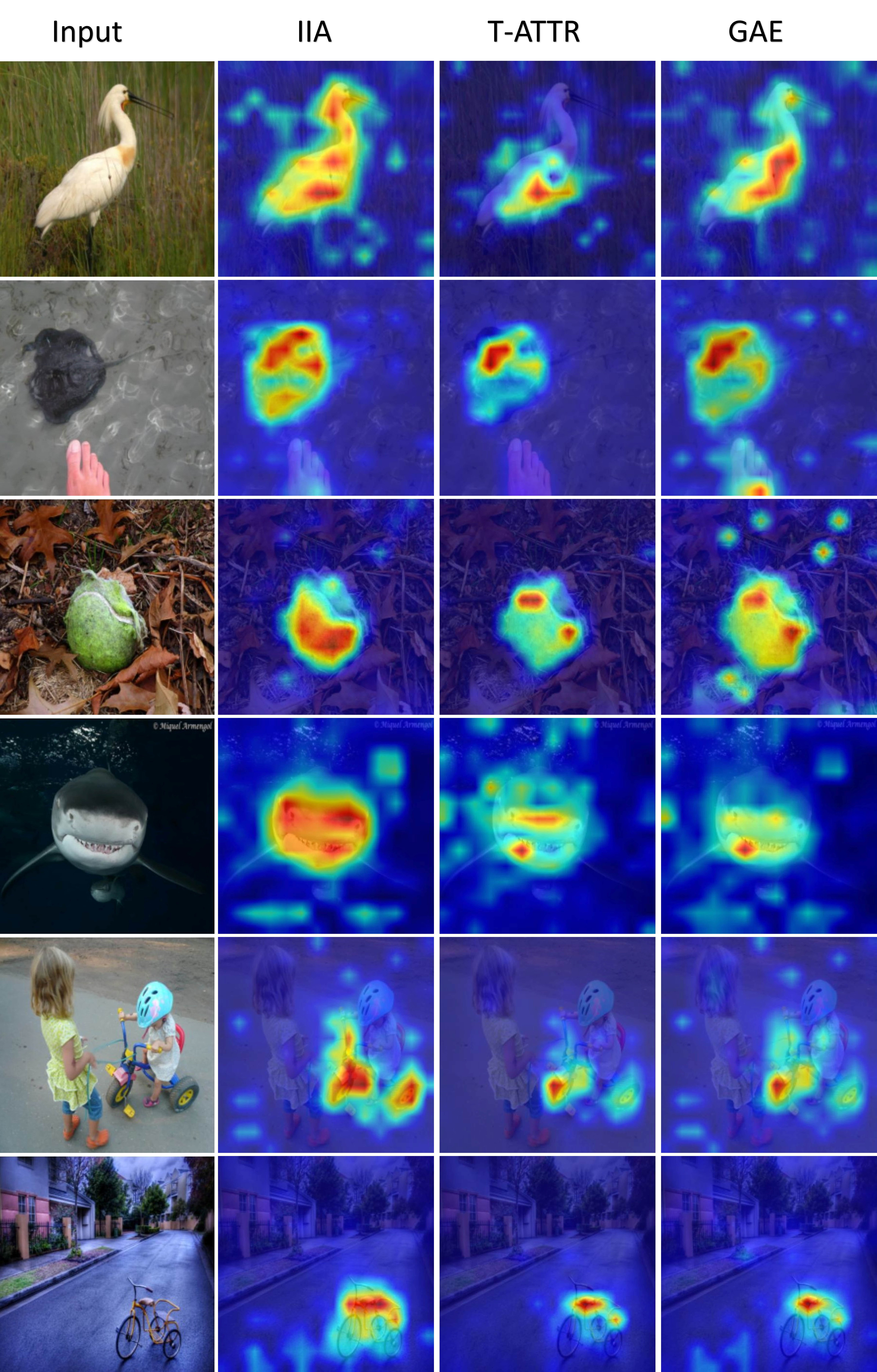}
    
    \caption{Visualizations obtained by explanation methods for ViT-B model. The ground-truth labels of the images are listed according to the format '($\langle$row\#$\rangle$) $\langle$class names$\rangle$': (1) 'spoonbill', (2) 'stingray', (3) 'tennis ball', (4) 'tiger shark, Galeocerdo cuvieri', (5-6) 'tricycle, trike, velocipede'.}
    \label{fig:qrv7}
\end{figure*}

\begin{figure*}
\centering
    \includegraphics[width=0.99\textwidth, height=0.95\textheight]{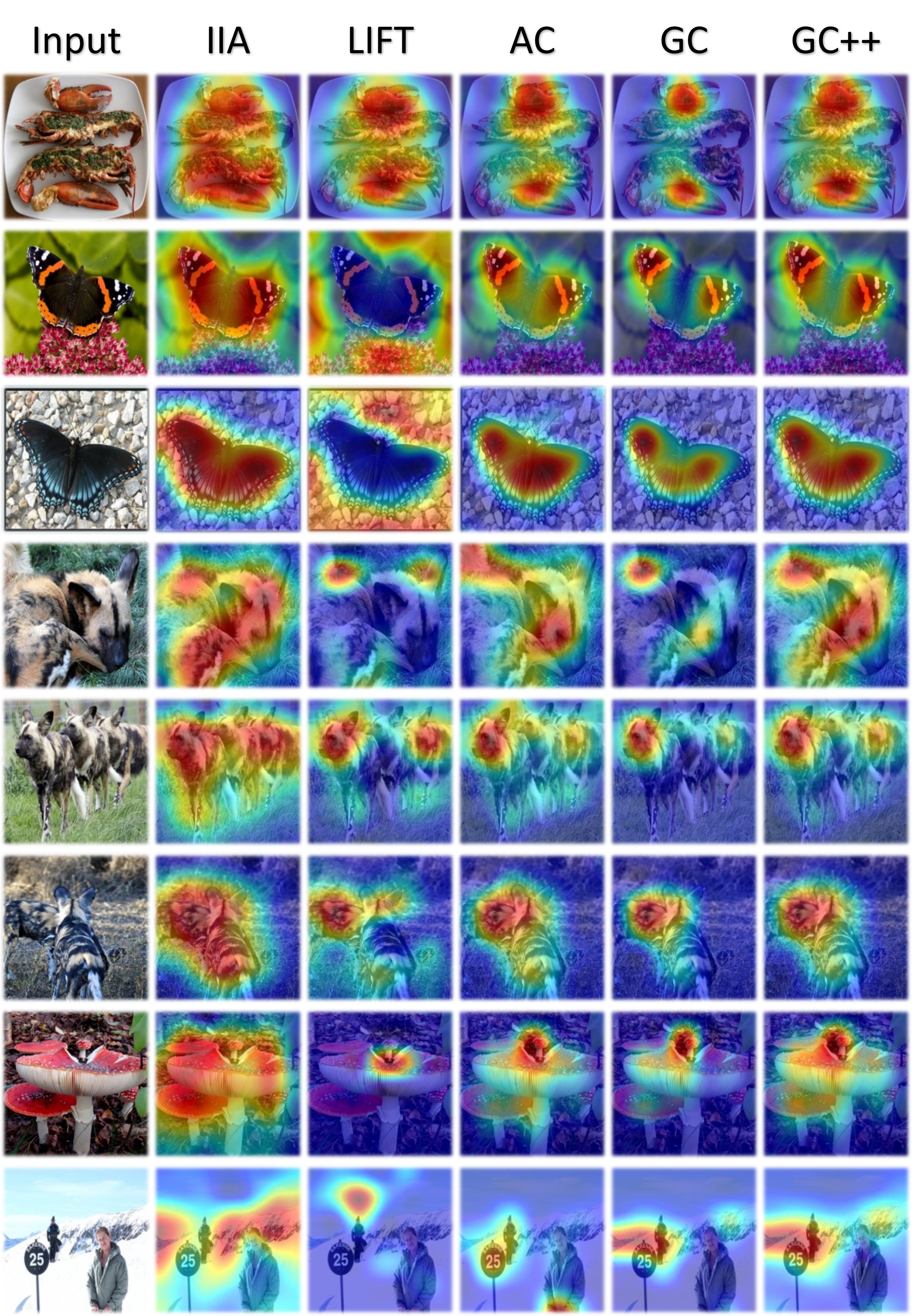}
    
    \caption{Visualizations obtained by the top performing methods in our evaluations. The ground-truth labels of the images are listed according to the format '($\langle$row\#$\rangle$) $\langle$class names$\rangle$': (1) 'American lobster, Northern lobster, Maine lobster, Homarus americanus', (2,3) 'admiral', (4-6) 'African hunting dog, hyena dog, Cape hunting dog, Lycaon pictus', (7) 'agaric', (8) 'alp'.  }
    \label{fig:qr1}
\end{figure*}

\begin{figure*}
\centering
    \includegraphics[width=0.99\textwidth, height=0.95\textheight]{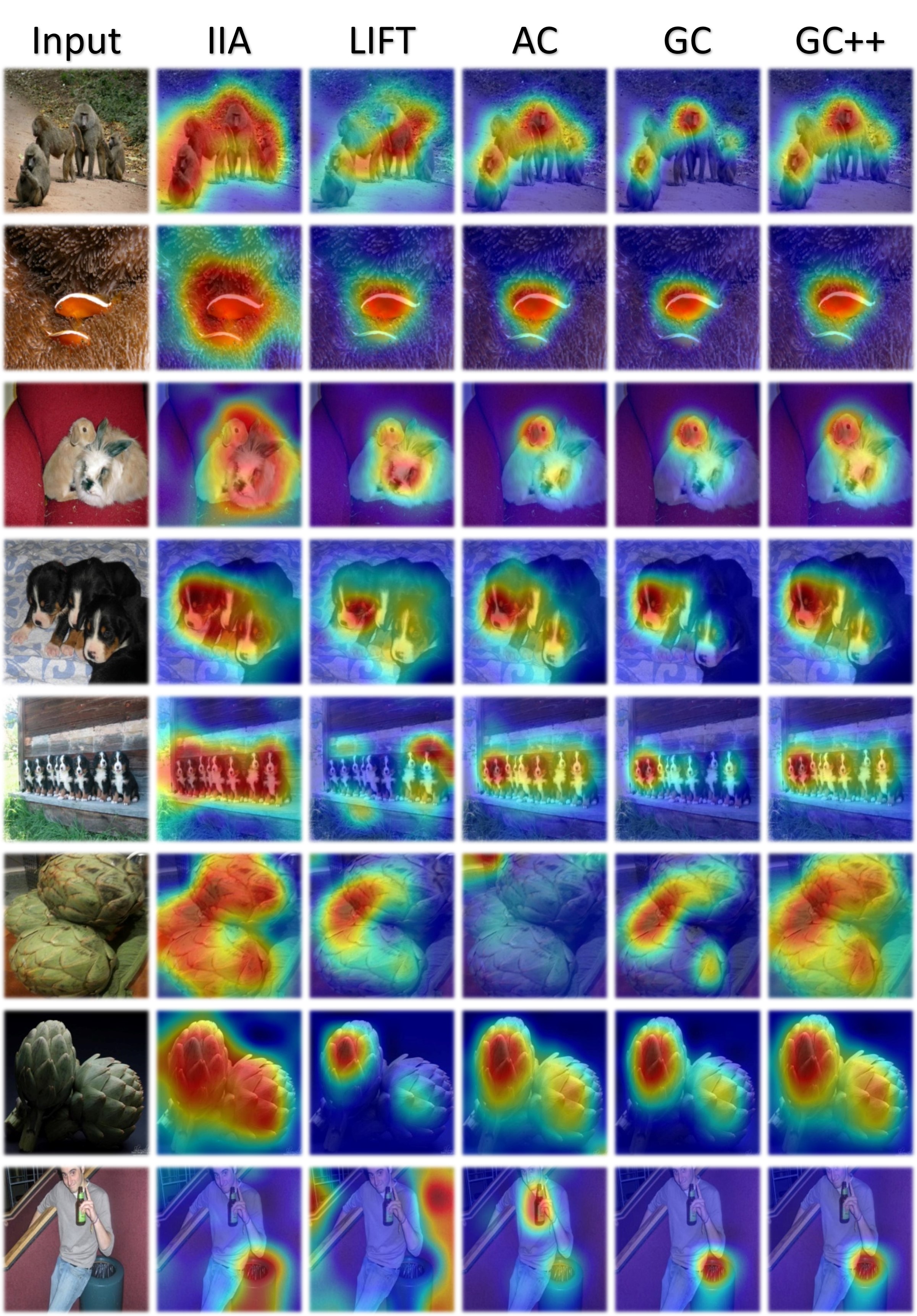}
    
    \caption{Visualizations obtained by the top performing methods in our evaluations. The ground-truth labels of the images are listed according to the format '($\langle$row\#$\rangle$) $\langle$class names$\rangle$': (1) 'baboon', (2) 'anemone fish', (3) 'Angora, Angora rabbit', (4,5) 'Appenzeller', (6,7) 'artichoke, globe artichoke', (8) 'ashcan, trash can, garbage can, wastebin, ash bin, ash-bin, ashbin, dustbin, trash barrel, trash bin'.}
    \label{fig:qr2}
\end{figure*}

\begin{figure*}
\centering
    \includegraphics[width=0.99\textwidth, height=0.95\textheight]{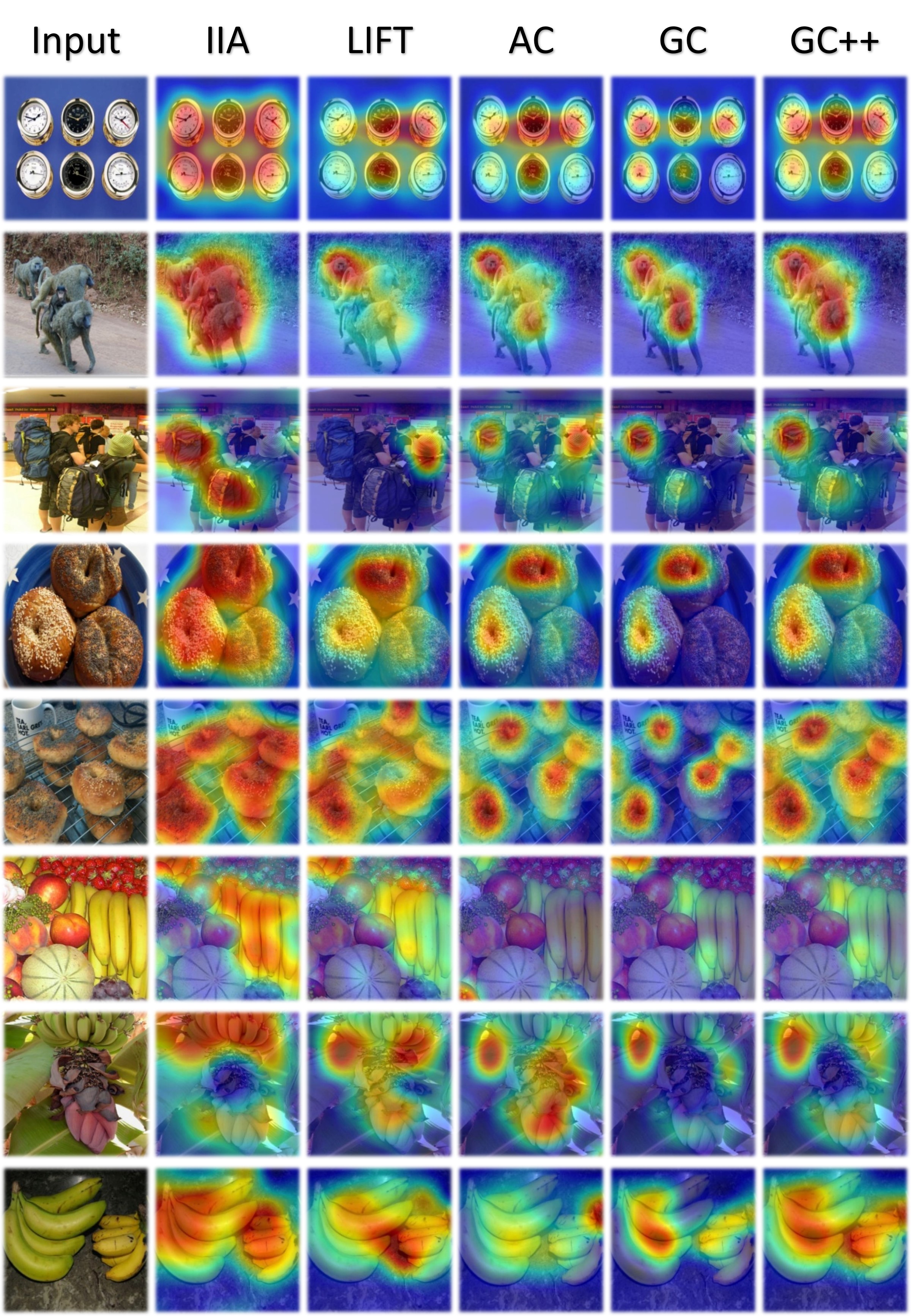}
    
    \caption{Visualizations obtained by the top performing methods in our evaluations. The ground-truth labels of the images are listed according to the format '($\langle$row\#$\rangle$) $\langle$class names$\rangle$': (1) 'barometer', (2) 'baboon', (3) 'backpack, back pack, knapsack, packsack, rucksack, haversack', (4,5) 'bagel, beigel', (6-8) 'banana'.}
    \label{fig:qr3}
\end{figure*}

\begin{figure*}
\centering
    \includegraphics[width=0.99\textwidth, height=0.95\textheight]{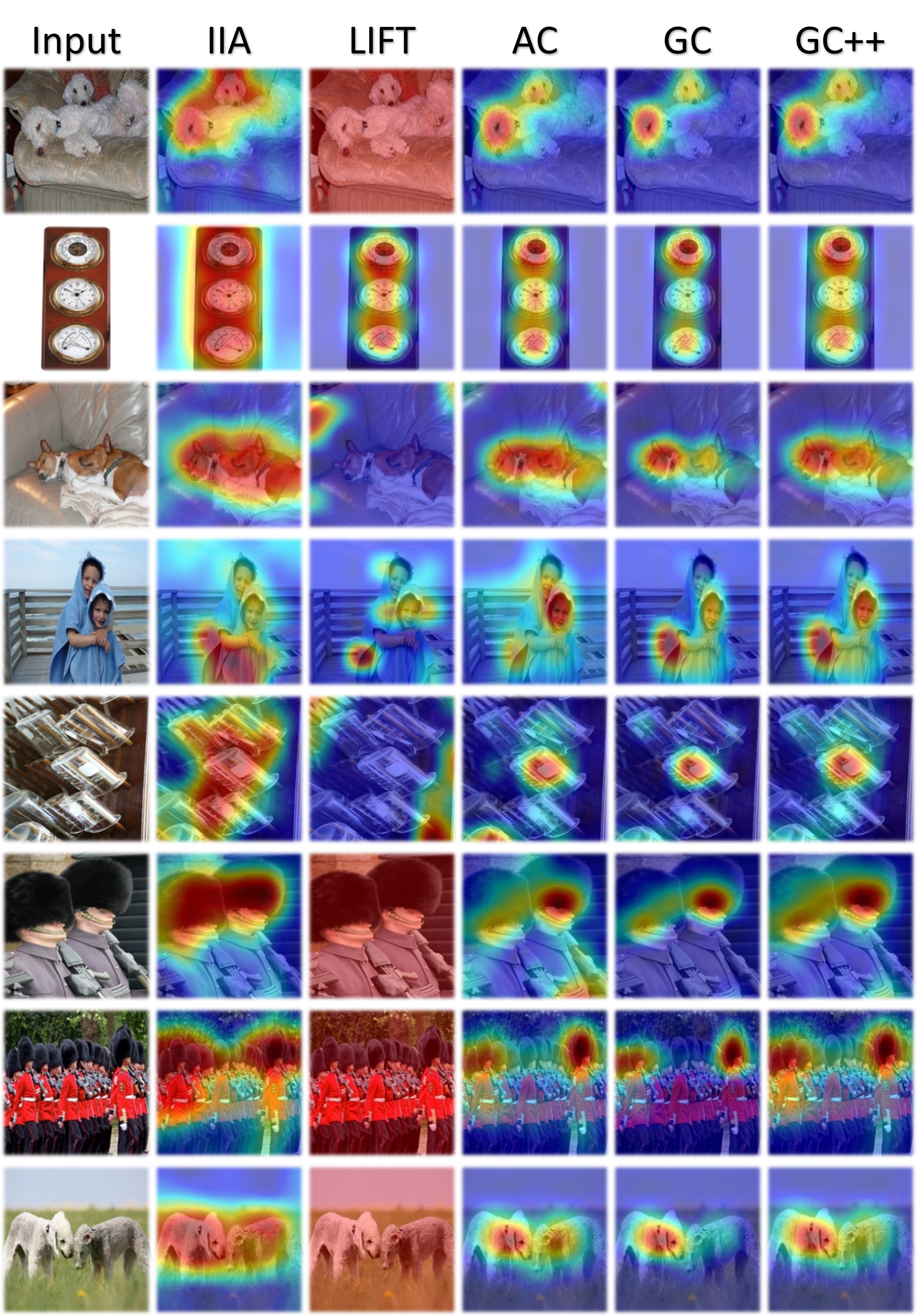}
    
    \caption{Visualizations obtained by the top performing methods in our evaluations. The ground-truth labels of the images are listed according to the format '($\langle$row\#$\rangle$) $\langle$class names$\rangle$': (1,8) 'Bedlington terrier', (2) 'barometer', (3) 'basenji', (4) 'bath towel', (5) 'beaker', (6,7): 'bearskin, busby, shako'. }
    \label{fig:qr4}
\end{figure*}

\begin{figure*}
\centering
    \includegraphics[width=0.99\textwidth, height=0.95\textheight]{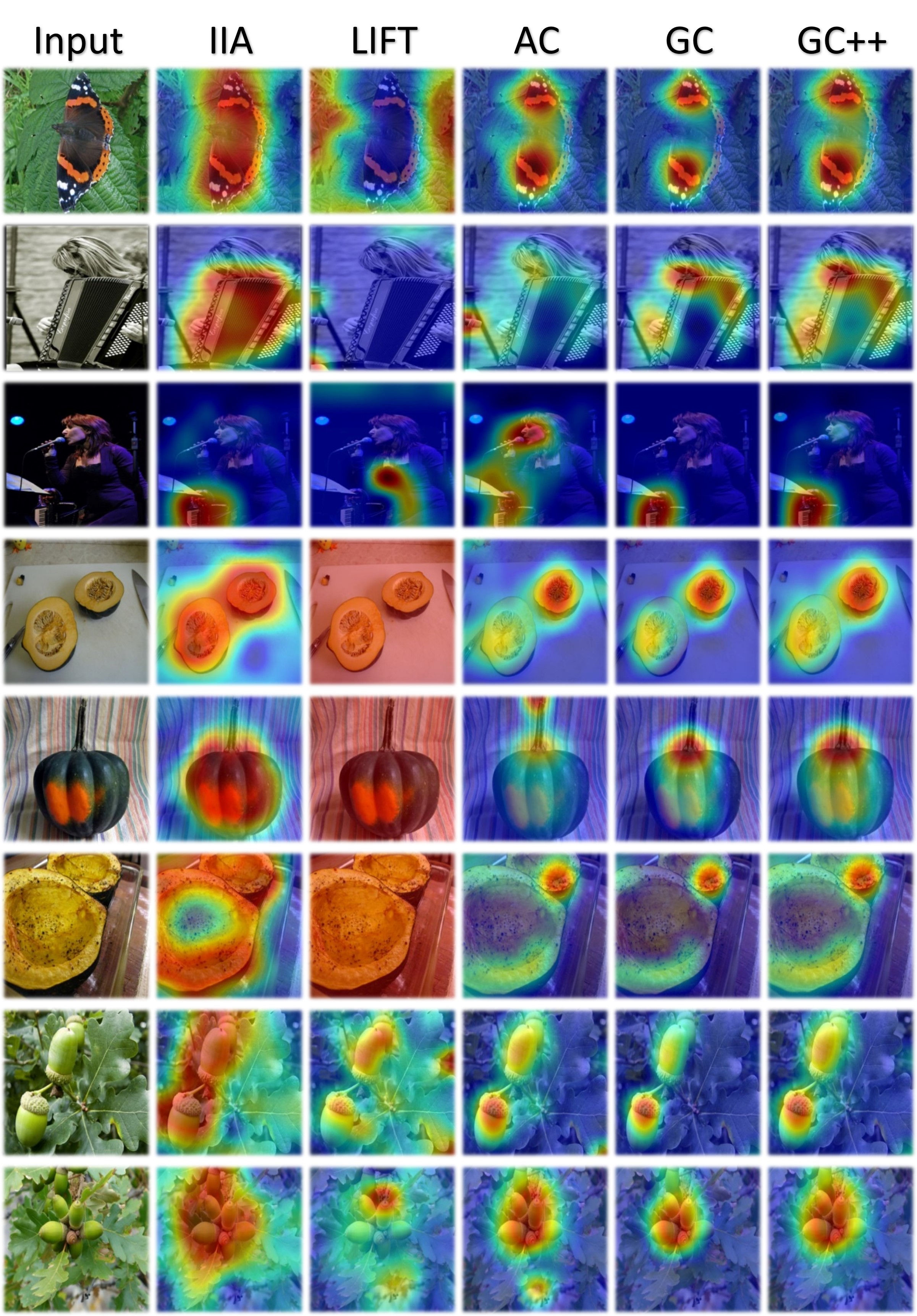}
    
    \caption{Visualizations obtained by the top performing methods in our evaluations. The ground-truth labels of the images are listed according to the format '($\langle$row\#$\rangle$) $\langle$class names$\rangle$': (1) 'admiral', (2,3) 'accordion, piano accordion, squeeze box', (4-6) 'acron squash', (7,8) 'acron'. }
    \label{fig:qr5}
\end{figure*}

\begin{figure*}
\centering
    \includegraphics[width=0.99\textwidth, height=0.95\textheight]{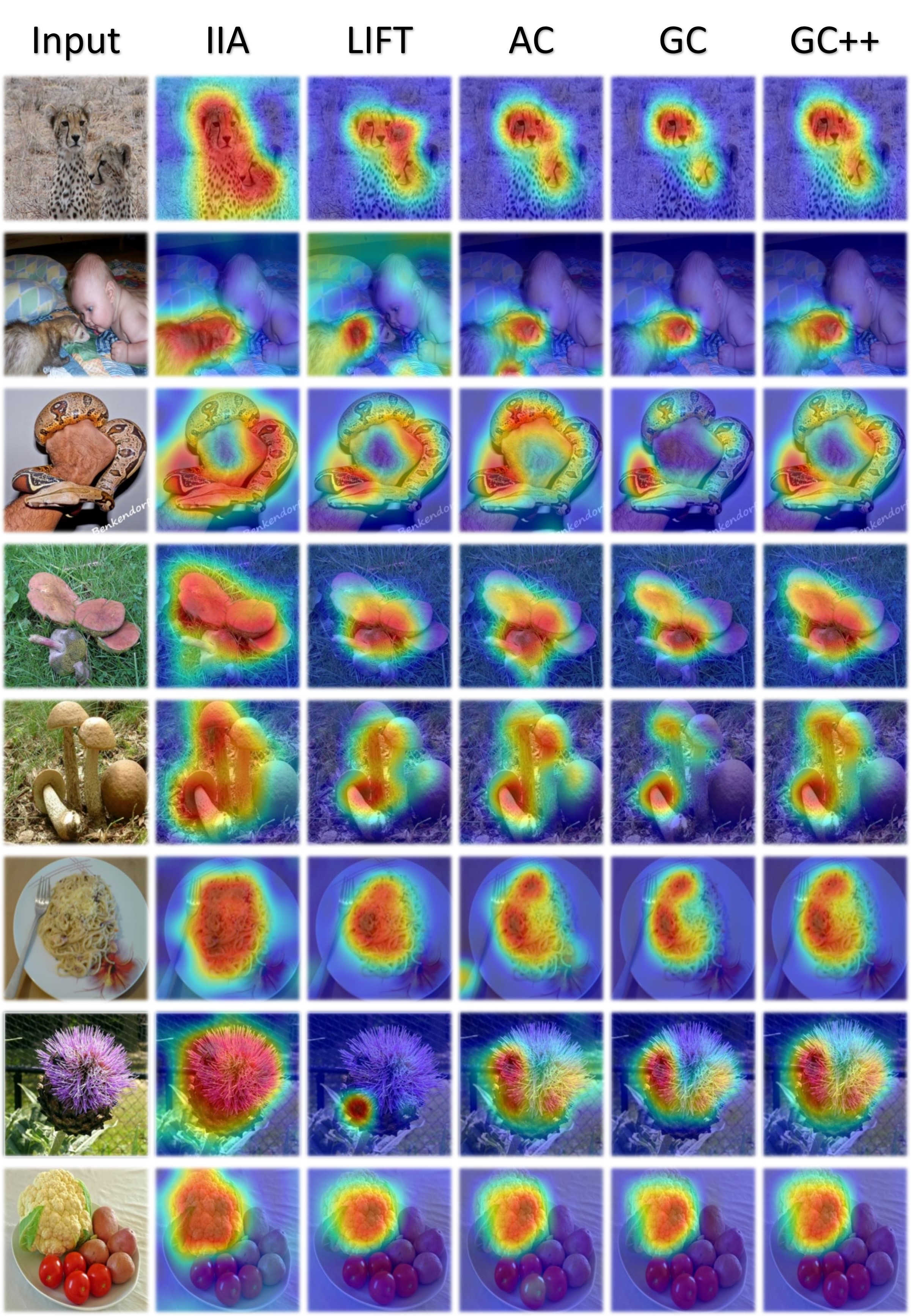}
    
    \caption{Visualizations obtained by the top performing methods in our evaluations. The ground-truth labels of the images are listed according to the format '($\langle$row\#$\rangle$) $\langle$class names$\rangle$': (1) 'cheetah, chetah, Acinonyx jubatus', (2) 'black-footed ferret, ferret, Mustela nigripes', (3) 'boa constrictor, Constrictor constrictor', (4,5) 'bolete', (6) 'carbonara', (7) 'cardoon', (8) 'cauliflower'.}
    \label{fig:qr6}
\end{figure*}

\begin{figure*}
\centering
    \includegraphics[width=0.99\textwidth, height=0.8\textheight]{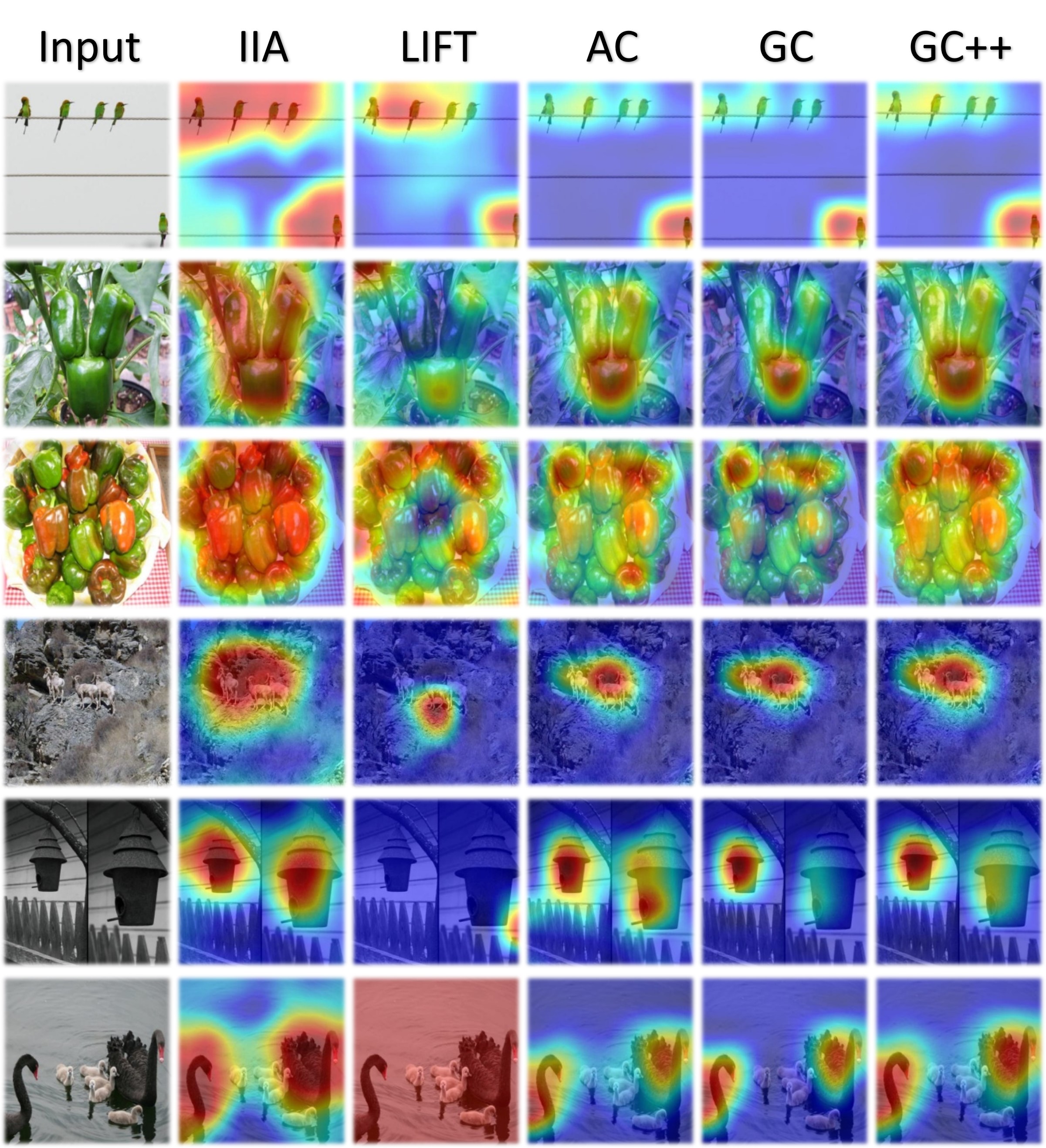}
    
    \caption{Visualizations obtained by the top performing methods in our evaluations. The ground-truth labels of the images are listed according to the format '($\langle$row\#$\rangle$) $\langle$class names$\rangle$': (1) 'bee eater', (2,3) 'bell pepper', (4) 'bighorn, bighorn sheep, cimarron, Rocky Mountain bighorn, Rocky Mountain sheep, Ovis canadensis', (5) 'birdhouse', (6) 'black swan, Cygnus atratus'.}
    \label{fig:qr8}
\end{figure*}

Figures~\ref{fig:qrv1}-\ref{fig:qrv7} present qualitative comparisons between our IIA method (IIA3), T-Attr~\cite{chefer2021transformer}, and GAE~\cite{chefer2021generic} (using the ViT-B model). Figures~\ref{fig:qr1}-\ref{fig:qr8} present qualitative comparisons between our IIA method (IIA3) and the best-performing methods from Tab.~\ref{tab:cnn_backbones_metrics_exp} (using the ConvNext model). The explanation maps are produced based on a random set of images sampled for various classes from the IN dataset. Arguably, IIA produces the most accurate explanation maps w.r.t. to the target classes both for CNNs and ViTs.

\end{document}